\documentclass[conference]{IEEEtran}

\usepackage{amsmath,amssymb,amsfonts}

\usepackage{algorithmic}

\usepackage{booktabs}
\usepackage{multirow}
\usepackage{diagbox}    % For diagonal headers
\usepackage{colortbl}   % For cell coloring
\usepackage[table]{xcolor} % Enhances table coloring

\usepackage{graphicx}
\usepackage{float}
\usepackage{subfigure}

\usepackage{textcomp}
\usepackage{xcolor}
\usepackage{etoolbox}   % Conditional macros
\usepackage{eso-pic}    % For adding background material
\usepackage{textpos}    % Absolute positioning of text

\usepackage{cite}
\usepackage{url}
\usepackage{hyperref}   % Should always be loaded last (except for cleveref, if used)

\usepackage{authblk}

\makeatletter
\renewcommand{\@cite}[2]{\textcolor{black}{[}\textcolor{green}{#1}\textcolor{black}{]}}
\makeatother

\newcommand{\myref}[1]{\textcolor{red}{\ref{#1}}}
\newcommand{\myeqref}[1]{\textcolor{red}{\eqref{#1}}}

\newcommand{\myhref}[2]{\textcolor{blue}{\href{#1}{#2}}}

\definecolor{matchRed}{RGB}{255,217,217}
\definecolor{matchGreen}{RGB}{217,255,217}
\definecolor{matchBlue}{RGB}{217,217,255}

\def\BibTeX{{\rm B\kern-.05em{\sc i\kern-.025em b}\kern-.08em
    T\kern-.1667em\lower.7ex\hbox{E}\kern-.125emX}}

\begin{document}

\title{Semantic Retention and Extreme Compression\\in LLMs: Can We Have Both?}

\author[*]{Stanislas Laborde}
\author[*]{Martin Cousseau}
\author[*]{Antoun Yaacoub}
\author[*]{Lionel Prevost}
\affil[*]{Learning, Data and Robotics (LDR) ESIEA Lab, ESIEA, Paris, France}

% Place the copyright notice in the bottom margin of the first page only
\AddToShipoutPictureFG*{% * means first page only
  \AtPageLowerLeft{% Position at the bottom left of the page
    \hspace{3em}% Adjust according to margins
    \raisebox{3em}{% Raise from the bottom of the page - adjust this value
      \makebox[\paperwidth-6em][c]{% Centered across the entire width (minus margins)
        \parbox{\paperwidth-6em}{\centering\footnotesize
          \textcopyright\ 2025 IEEE. Personal use of this material is permitted. Permission from IEEE must be obtained for all other uses, in any current or future media.\\[0.5em]
          Accepted for publication in the \textit{Proceedings of the 2025 International Joint Conference on Neural Networks (IJCNN)}.
        }
      }
    }
  }
}

\maketitle

\begin{abstract}
The exponential growth in Large Language Model (LLM) deployment has intensified the need for efficient model compression techniques to reduce computational and memory costs. While pruning and quantization have shown promise, their combined potential remains largely unexplored. In this paper, we examine joint compression and how strategically combining pruning and quantization could yield superior performance-to-compression ratios compared to single-method approaches. Recognizing the challenges in accurately assessing LLM performance, we address key limitations of previous evaluation frameworks and introduce the Semantic Retention Compression Rate (\textit{SrCr}), a novel metric that quantifies the trade-off between model compression and semantic preservation, facilitating the optimization of pruning-quantization configurations. Experiments demonstrate that our recommended combination achieves, on average, a 20\% performance increase compared to an equivalent quantization-only model at the same theoretical compression rate.
\end{abstract}


\vspace{-1em}

\section{Introduction} \label{introduction}
Large Language Models (LLMs) have demonstrated remarkable capabilities across a wide range of tasks, from natural language understanding to complex reasoning. However, their increasing size---from GPT-3's 175 billion parameters to GPT-4's estimated trillions---presents significant deployment challenges, particularly in resource-constrained environments.

Recent years have seen the development of various compression approaches, which are usually either training-aware or post-training. These approaches include low-rank matrix factorization, which reduces parameter count by decomposing weight matrices~\cite{lora}; knowledge distillation, which transfers knowledge from larger to smaller models~\cite{distilbert}; pruning, which removes less important connections~\cite{wanda}; and quantization, which reduces the precision of model weights~\cite{gptq}. The field has evolved from modest to aggressive techniques, with quantization being pushed to its limits through 1-bit representations~\cite{quant1bit}, though such extremes often struggle to preserve model performance, particularly on complex reasoning tasks.

Complementary to parameter-focused compression are graph optimization techniques (e.g., ONNX, TensorRT), which improve inference efficiency through computational graph transformations without directly modifying model parameters.

The challenge of evaluating compressed models adds another layer of complexity. Traditional metrics like perplexity have been shown to correlate poorly with actual model capabilities, especially under compression. As demonstrated by~\cite{llmkick}, perplexity often fails to capture subtle degradation in knowledge-intensive tasks even when compressed models maintain similar perplexity scores to their dense counterparts. Moreover, recent studies have shown an inverse relationship between model size and information density, with larger models often having lower parameter efficiency~\cite{densinglaw}. This phenomenon manifests in compression experiments, where larger models demonstrate disproportionately high compression rates that may not generalize to more efficient architectures. This can lead to overstated claims, as seen with techniques like SparseGPT reporting extreme pruning rates primarily based on perplexity metrics~\cite{sparsegpt}, potentially misrepresenting their general applicability. Notably, empirical evidence suggests model capability density doubles every three months through algorithmic improvements~\cite{densinglaw}, indicating a clear trend toward more efficient architectures and training methodologies.

Rather than pursuing ever more extreme forms of standalone compression, a promising direction may lie in combining multiple approaches. Yet, despite the potential synergies between methods like pruning and quantization, their joint application for LLMs has remained surprisingly underexplored. The few existing attempts at joint compression have been limited in scope~\cite{sparsegpt} or focused on specific architectures~\cite{jsq}, leaving open questions regarding the broader potential of this approach.

In this work, we address these challenges through three main contributions:

(1) We develop a theoretical framework to analyze joint compression, introducing the concept of Theoretical Compression Rate ($TCr$) to enable fair comparisons across setups.

(2) We propose novel metrics to quantify semantic retention in compressed models, providing a principled way to evaluate the trade-off between model size and capability preservation.

(3) We conduct extensive experiments across compression setups, including unstructured and semi-structured approaches, revealing optimal joint compression points that significantly outperform single-method approaches at the same $TCr$ values.\\

The remainder of this paper is organized as follows. Section~\myref{background} reviews background on compression techniques and related work. Section~\myref{methodology} presents our methodology, including the theoretical framework, sequential approximation, and novel semantic retention metrics. Section~\myref{experiments} details our experiments and compares various compression configurations. Section~\myref{conclusion} concludes our findings, while Section~\myref{limitations} discusses limitations and outlines future directions, including unified joint compression algorithms and hardware-aware optimization strategies.

\section{Background} \label{background}
\subsection{Pruning}
While training-aware pruning traditionally achieved better results, its computational demands proved impractical for billion-parameter models~\cite{wanda}. This led to efficient post-training approaches like SparseGPT~\cite{sparsegpt}, which enables one-shot pruning through sparse regression. Unstructured pruning offers maximum theoretical compression by removing individual weights, but often results in irregular sparsity patterns that are challenging to accelerate on current hardware. Semi-structured approaches, like N:M sparsity patterns~\cite{2021NM}, balance compression rates with hardware efficiency by enforcing regular pruning patterns, where N out of every M consecutive weights are pruned. Structured pruning~\cite{llmpruner} takes this further by removing entire structures, channels, or attention heads, with recent work showing that up to 50\% of attention layers in large models can be removed while preserving performance~\cite{whatmatters}.

\subsection{Post-Training Quantization}
Recent advances in quantization have demonstrated impressive results in maintaining model performance while significantly reducing precision requirements. LLM.int8()~\cite{llmint8} established 8-bit quantization using outlier-aware techniques, while GPTQ~\cite{gptq} pushed boundaries with 4-bit quantization through second-order optimization. AWQ~\cite{AWQ} introduced activation-aware weight quantization that more effectively preserves model capabilities, while SmoothQuant~\cite{SmoothQuant} proposed effective weight-activation redistribution. QuantEase~\cite{QuantEase} developed a cyclic coordinate descent approach that achieves state-of-the-art accuracy in 4-bit and 3-bit quantization while remaining highly memory-efficient. OmniQuant~\cite{OmniQuant} combined learnable weight clipping with equivalent transformations to enable efficient low-bit quantization. Additional innovations include channel-wise quantization approaches~\cite{channel-wise},~\cite{adadim}, and advances in outlier handling~\cite{OWQ}.

\subsection{Related Work}
Recent research in LLM compression has pursued several parallel tracks. Some approaches focus on combining multiple compression methods, like exploring fine-grained quantization with pruning~\cite{spqr}, or sensitivity-based quantization with sparse-dense decomposition~\cite{squeezellm}. The development of hardware-efficient formats has led to innovations in structured sparsity patterns~\cite{sdq} and unified compression frameworks~\cite{jsq}.

In contrast, our work differs from these implementation-focused approaches by establishing a theoretical framework for understanding and comparing compression techniques. Rather than proposing another specific compression technique, we develop principled metrics and analysis tools applicable across different compression configurations. This theoretical foundation provides insights complementing practical approaches and enabling more rigorous evaluation of compression effectiveness.

\section{Methodology} \label{methodology}
\subsection{Theoretical Framework for Joint Compression}
To systematically investigate joint compression, we first establish a theoretical framework for comparing different compression techniques.

\medskip
\subsubsection{Theoretical Compression Rate}
We introduce the concept of Theoretical Compression Rate ($TCr$), which measures the fundamental compression of information in the model, independent of implementation-specific overheads. For a given compression configuration, $TCr$ is defined as:
\begin{equation}
    TCr = \begin{cases}
        \frac{s}{100} & \text{pruning-only} \\
        1 - \frac{q}{16} & \text{quantization-only} \\
        1 - \frac{q}{16} \frac{100-s}{100} & \text{joint compression}
    \end{cases}
\end{equation}
where $s$ represents the sparsity percentage and $q$ denotes the quantization bit-width. We use 16-bit as the baseline since modern LLMs are typically distributed in this precision without performance loss, making it the practical reference for compression. While defined mathematically as a ratio for conciseness and readability, $TCr$ values are hereafter expressed as percentages.

\medskip
\subsubsection{Properties of Theoretical Compression Rate}
We prioritize $TCr$ over physical compression rates for three key reasons:

\begin{itemize}
    \item It provides a fundamental measure of information compression, independent of non-informative storage requirements like pruning masks, or negligible overheads such as dequantization weights;
    \item It enables fair comparison across compression approaches: between unstructured and semi-structured pruning at identical sparsity levels (as shown in our experiments), and potentially between pruning and/or quantization techniques, despite their differing overheads;
    \item It abstracts hardware-specific considerations like throughput optimization, which remain active research areas.
\end{itemize}

\subsection{Sequential Approximation Approaches}
To evaluate the potential of joint compression, we argue that it can be approximated through the sequential application of pruning followed by quantization.

\medskip
\subsubsection{Sequential vs. Joint Compression}
Information lost during compression cannot be recovered without fine-tuning. When pruning is applied at a sparsity rate $s$, the model's knowledge is compressed into the remaining $(100-s)\%$ weights, with the pruned weights becoming informationally-void. Similarly, quantization compresses the precision and range of weights, and any information lost during this process cannot be recovered through dequantization alone.

In an optimal joint compression setting, pruning and quantization would work synergistically, with each operation considering the other's impact on the model's information content. However, in sequential compression, these operations work independently, potentially leading to increased error accumulation.

Therefore, we can justify sequential compression as an approximation of joint compression through two key \mbox{observations}:

\begin{itemize}
    \item The fundamental irreversibility of information loss means that the order of operations primarily affects error distribution rather than total information content;
    \item While sequential application may not achieve optimal error distribution, it preserves the core compression characteristics of both operations.
\end{itemize}

\medskip
\subsubsection{Analysis of Approximation Approaches}
We analyze a potential approximate approach to sequential compression: quantizing all weights after pruning (Case A) rather than only the non-pruned weights (Case B). While Case B represents the more intuitive implementation, we examine how Case A might serve as a viable approximation, particularly at lower sparsity levels, and analyze the theoretical foundations supporting this possibility. Consider the GPTQ objective:
\begin{equation}
    \operatorname*{argmin}_{\widehat{W}} \|WX - \widehat{W}X\|_2^2
\end{equation}
where $W$ is the weights matrix, $\widehat{W}$ is the quantized weights matrix, and $X$ is the input matrix (calibration data).

For Case A (quantizing all weights) and Case B (quantizing only non-pruned weights), we analyze the error introduced by quantization. Let $M$ be the sparsity mask where $M_{i,j} = 1$ if $W_{i,j} \neq 0$ and 0 otherwise. The column-wise errors are:
\begin{align}
    E_{A,j}^{GPTQ} &= \|w_{A,j} - q_{A,j} + \delta_{A,j}\|_2^2 \label{eq:error_a} \\
    E_{B,j}^{GPTQ} &= \|M_{:,j} \odot (w_{B,j} - q_{B,j}) + \delta_{B,j}\|_2^2 \label{eq:error_b}
\end{align}
where $w_{A,j}$ and $w_{B,j}$ represent column $j$ of the continuously updated weight matrix for cases A and B respectively, $q_{A,j}$ and $q_{B,j}$ their corresponding quantized versions, and the accumulated updates $\delta_{A,j}$ and $\delta_{B,j}$ are given by:
\begin{align}
    \delta_{A,j} &= -\sum_{k<j} (w_{A,k} - q_{A,k}) \cdot [H^{-1}]_{k,j} \label{eq:delta_a} \\
    \delta_{B,j} &= -\sum_{k<j} (M_{:,k} \odot (w_{B,k} - q_{B,k})) \cdot [H_M^{-1}]_{k,j} \label{eq:delta_b}
\end{align}
where $H$ is the Hessian matrix and $H_M$ represents the Hessian matrix that inherently respects the sparsity pattern of the pruned weights. Both the weight matrices and Hessian matrices are updated after processing each block of 128 columns, allowing the algorithm to maintain accurate second-order information throughout the quantization process.

These error terms can be understood as the combination of two distinct components. First, there is an immediate quantization error arising from the loss of precision between the current updated weights and their quantized values ($w_{A,j} - q_{A,j}$ and $M_{:,j} \odot (w_{B,j} - q_{B,j})$ terms). Second, there is a cumulative error component ($\delta_{A,j}$ and $\delta_{B,j}$ terms) that propagates from the quantization of previous columns through the Hessian-based updates. This dual error structure is fundamental to understanding the behavior of GPTQ in both cases.

The error ratio analysis shows:
\begin{equation}
    \frac{E_{A,j}^{GPTQ}}{E_{B,j}^{GPTQ}} \approx \frac{n\varepsilon_q^2 + \|\delta_{A,j}\|_2^2}{(1-p)n\varepsilon_q^2 + \|\delta_{B,j}\|_2^2} \label{eq:ratio}
\end{equation}
where $p=\frac{s}{100}$ is the pruning ratio, $n$ is the dimension of the weight vectors, and $\varepsilon_q$ represents the statistical average quantization error when quantizing a random 16-bit weight to the target precision $q$.

The error ratio analysis reveals a complementary trade-off between the two approaches:
\begin{itemize}
    \item Case A provides more degrees of freedom for error correction by allowing updates to all weights and using full Hessian information, but introduces additional error through updates to pruned weights and potentially overcompensating correlations;
    \item Case B maintains a smaller error footprint through masked Hessian correlations and restricted updates to non-pruned weights, but has fewer weights available for error distribution.
\end{itemize}

The ratio $\frac{E_{A,j}^{GPTQ}}{E_{B,j}^{GPTQ}}$ approaches 1 when $p$ tends toward zero, as the mask M approaches the identity matrix, indicating that these competing effects become negligible at lower sparsity levels. This analysis, combined with the principle that information loss in compression is fundamentally irreversible, supports using Case A as a practical approximation for studying joint compression, at least for small sparsity levels. We adopt this approach for several reasons:
\begin{itemize}
    \item Implementation-agnostic: Case A allows the evaluation of different pruning and quantization techniques without consideration of specific sparsity patterns or quantization algorithms;
    \item Experimental flexibility: Case A facilitates direct comparison between joint compression and single-method configurations;
    \item Implementation efficiency: Case A enables broader exploration of compression configurations without requiring specialized implementations for each technique combination.
\end{itemize}

This choice aligns with our goal of understanding the fundamental potential of joint compression, while acknowledging that optimal implementations may eventually require more sophisticated approaches.

\medskip
\subsubsection{Validation of Error Approximation}
To validate our theoretical analysis, we leverage simpler quantization techniques (LLM.int8()~\cite{llmint8}, NF4~\cite{qlora}) whose error terms are analytically tractable. For these techniques, the column-wise errors in Case~A~are:
\begin{equation}
E_{A,j}^{simple} = n\varepsilon_q^2 \label{eq:error_simple} 
\end{equation}

The difference in performance degradation between GPTQ and simpler techniques after pruning directly reflects the impact of the accumulated error term $\|\delta_{A,j}\|_2^2$, which approaches $\|\delta_{B,j}\|_2^2$ when $p$ tends toward zero. Specifically:
\begin{equation}
E_{A,j}^{GPTQ} - E_{A,j}^{simple} \approx \|\delta_{A,j}\|_2^2 \label{eq:error_a_simple}
\end{equation}

This relationship allows us to empirically corroborate our theoretical framework by comparing performance differences between GPTQ and simpler techniques (NF4 at 4-bit, LLM.int8() at 8-bit) before and after pruning (25\% and 50\% respectively). If these differences remain small, it provides evidence that $\|\delta_{A,j}\|_2^2$ is well-bounded, supporting our use of Case A as an approximation for Case B in studying joint compression.

\subsection{Retention Metrics}
To systematically evaluate compression quality, we need metrics that capture both task-specific performance degradation and overall semantic preservation. While compression rates provide a measure of model size reduction, they don't directly reflect the preservation of model capabilities. We therefore introduce retention metrics that quantify how well a compressed model maintains the performance of its original counterpart across different tasks and semantic domains.

We measure the retention rate $R_t$ for a compressed model $M_c$ relative to its original model $M_o$ on a task $t$ as:
\begin{equation}
    R_t = \frac{P_{M_c}(t)}{P_{M_o}(t)}
\end{equation}
where $P_M(t)$ denotes the performance of model $M$ on task $t$.

\medskip
\subsubsection{Semantic Retention Rate}
The Semantic Retention Rate ($Sr$) extends the task-specific retention metric to capture the overall preservation of model capabilities across the entire task space. This generalization yields two natural formulations, each offering different perspectives on how to aggregate retention across tasks:
\vspace{-0.4em}
\begin{equation}
    \hspace{-5.8em}Sr_1 = \lim_{n \rightarrow \infty} \frac{1}{n} \sum_{i=1}^n \frac{P_{M_c}(t_i)}{P_{M_o}(t_i)} = \lim_{n \rightarrow \infty} \frac{1}{n} \sum_{i=1}^n R_{t_i} \\
\end{equation}
\vspace{-1.375em}
\begin{equation}
\hspace{-0.3em}Sr_2 = \lim_{n \rightarrow \infty} \frac{\sum_{i=1}^n P_{M_c}(t_i)}{\sum_{i=1}^n P_{M_o}(t_i)} = \lim_{n \rightarrow \infty} \sum_{i=1}^n \frac{P_{M_o}(t_i) R_{t_i}}{\sum_{j=1}^n P_{M_o}(t_j)}
\end{equation}
where $\{t_i\}_{i=1}^n$ represents a set of $n$ independent evaluation tasks. $Sr_1$ weights each retention ratio equally, while $Sr_2$ weights by original performance magnitude. In simpler terms, $Sr_1$ represents the average of retentions across all tasks, while $Sr_2$ represents the retention of the average performance across all tasks. Since tasks are inherently hierarchical (composed of sub-tasks) and task performances already represent averages over multiple instances, $Sr_2$ provides a more natural aggregation across these nested evaluation structures. We therefore adopt $Sr_2$ as our semantic retention metric for the remainder of this work and refer to it simply as $Sr$ hereafter.

In practice, this theoretical rate is approximated over a finite set of tasks $\mathcal{T}$ that should maximize coverage of the model's capability space while maintaining task independence. The selection of $\mathcal{T}$ should balance breadth (diverse task types), depth (multiple difficulty levels), and orthogonality (minimizing task overlap) to provide meaningful retention estimates.

\medskip
\subsubsection{Semantic Retention Compression Rate}
When comparing models compressed at different $TCr$ values, we would like a metric that captures both the degree of compression and the preservation of model capabilities. This metric should:

\begin{itemize}
    \item Increase with better performance retention;
    \item Take account of compression aggressiveness;
    \item Capture known empirical behaviors of compression \mbox{methods};
    \item Be bounded and interpretable.
\end{itemize}

These first two criteria exhibit an inherent trade-off: as compression becomes more aggressive, model performance typically deteriorates. Our objective is to identify a maximum in this relationship, which we hypothesize occurs at the theoretically optimal configuration for joint compression.

We develop separate formulations for pruning and quantization before combining them into a unified metric.

\smallskip
\paragraph{Pruning Component}
For pruning, although memory savings are directly proportional to the pruning ratio $p$, empirical evidence demonstrates that using this linear term in our metric would incorrectly favor high sparsity levels. We instead adopt a formulation that emphasizes practically-relevant sparsity ranges, with the square root of $p$ providing appropriate sensitivity to compression levels:
\begin{equation}
SrCr_{p} = \sqrt{p} \cdot Sr_p \label{eq:srcr_p}
\end{equation}
where $Sr_p$ is the semantic retention rate. The square root function naturally normalizes the metric to $[0,1]$ while providing higher resolution in the critical lower pruning ranges, better reflecting the trade-off between compression and capability retention. This formulation:
\begin{itemize}
    \item Provides greater sensitivity in the practically-viable \mbox{pruning} range $[0\%,50\%]$, beyond which models show significant degradation;
    \item Reflects empirical observations of performance deterioration at high sparsity levels.
\end{itemize}

\smallskip
\paragraph{Quantization Component}
For quantization, we adopt an information-theoretic perspective. The reduction in bit-width represents a logarithmic decrease in representational capacity:
\begin{equation}
    SrCr_q = \frac{-\log_2(q/16)}{4} \cdot Sr_q \label{eq:srcr_q}
\end{equation}
where $q$ is the quantization bit-width and $Sr_q$ is the semantic retention rate. The denominator $4$ corresponds to the reference point at 1-bit quantization, representing the maximum logarithmic compression achievable ($\log_2(16/1) = 4$). This formulation:
\begin{itemize}
    \item Reflects the exponential relationship between bits and precision;
    \item Matches empirical observations of sharp performance drops below critical bit-widths.
\end{itemize}

\smallskip
\paragraph{Joint Metric}
Combining these components multiplicatively yields our joint compression metric:
\begin{equation}
    SrCr_{j} = \left[\frac{-\log_2(q/16)}{4} \cdot \sqrt{p}\right] \cdot Sr_j \label{eq:srcr_j}
\end{equation}

This normalization approach ensures all resulting metrics have several desirable properties:
\begin{itemize}
    \item Bounded range $[0,1]$ through division by theoretical maxima;
    \item Maximum value of 1 occurs at optimal configurations with perfect retention;
    \item Minimum value of 0 occurs at complete information loss or no joint configuration $(p,q)$;
    \item Smooth gradients for optimization;
    \item Interpretability and fair comparison across configurations.
\end{itemize}

\section{Experiments} \label{experiments}
\subsection{Experimental Settings}
As discussed in Section~\myref{introduction}, the relationship between model size and information density poses key challenges for compression research. Building upon these insights, we focus on smaller, information-dense LLMs rather than larger, potentially overparameterized models. While our theoretical framework is model-agnostic, we emphasize the importance of developing algorithms that scale efficiently with model size \mbox{---such} as layer-wise approaches like SparseGPT and GPTQ--- to avoid exponentially growing computational complexity.

For our experiments, we selected two state-of-the-art open-source language models that maintain parameter efficiency:
\begin{itemize}
    \item \myhref{https://huggingface.co/meta-llama/Llama-3.1-8B}{LLaMA-3.1-8B}: A multilingual model with robust performance across diverse tasks~\cite{llama3};
    \item \myhref{https://huggingface.co/mistralai/Mistral-7B-v0.3}{Mistral-7B-v0.3}: A monolingual model optimized for English-language tasks~\cite{mistral7b}.
\end{itemize}

We chose SparseGPT and GPTQ as compression algorithms for their top performance and public implementations (from \myhref{https://github.com/vllm-project/llm-compressor}{\texttt{llmcompressor}}) supporting various bit-widths and pruning patterns, enabling comprehensive comparative analysis.

\medskip
\subsubsection{Benchmark Selection}
Following careful analysis of recent evaluation frameworks and their limitations~\cite{survey1}\textcolor{green}{--}\cite{hf}, we selected three complementary, robust benchmarks that address previous issues of dataset contamination and quality while emphasizing different aspects of model capabilities:%\cite{survey1, survey2, hf}

\begin{itemize}
    \item MMLU-Pro~\cite{mmlupro}: An enhanced version of MMLU that evaluates knowledge breadth and depth across 14 academic disciplines. Key improvements include:
    \begin{itemize}
        \item Expanded multiple choice format (10 vs. 4 options);
        \item Enhanced focus on reasoning over pure knowledge retrieval;
        \item Expert review process for question quality assurance;
        \item Systematic noise reduction through verification protocols.
    \end{itemize}
    
    \item BBH (Big-Bench Hard)~\cite{bbh}: A curated subset of 23 challenging tasks from BIG-Bench, selected for:
    \begin{itemize}
        \item Performance below the human baseline in prior \mbox{evaluations};
        \item Emphasis on multi-step reasoning capabilities;
        \item Objective evaluation metrics with statistical significance;
        \item Strong correlation with human preference metrics.
    \end{itemize}
    
    \item MATH~\cite{MATH}: A collection of competition-level mathematics problems spanning multiple domains, with:
    \begin{itemize}
        \item High-difficulty questions from mathematics competitions;
        \item Standardized \LaTeX{} formatting for equations;
        \item Required step-by-step solution documentation;
        \item Precise answer format specifications.
    \end{itemize}
\end{itemize}

While the \myhref{https://huggingface.co/spaces/open-llm-leaderboard/open_llm_leaderboard\#}{HuggingFace Open LLM Leaderboard v2} includes other notable benchmarks (IFEval~\cite{ifeval}, MuSR~\cite{musr}, and GPQA~\cite{gpqa}), we excluded these to minimize task overlap and redundancy in our evaluation criteria. This decision was further supported by preliminary testing indicating that our baseline 7--8B models showed insufficient performance on some of these tasks for meaningful compression analysis.

\medskip
\subsubsection{Configuration of Compression Techniques}
We applied SparseGPT and GPTQ on all transformer layers with the following configurations:
\begin{itemize}
\item SparseGPT:
\begin{itemize}
    \item Dataset: C4 (default), 512 calibration samples;
    \item Block size: 128 (default);
    \item \textit{Dampening} parameter: 0.01 (default).
\end{itemize}
\item GPTQ:
\begin{itemize}
    \item Dataset: C4 (default);
    \item Group size: 128 (default).
\end{itemize}
\end{itemize}

\medskip
\subsubsection{Implementation Details}
All experiments were conducted on NVIDIA A10 GPUs with 24GB of memory. For standardized evaluation, the \myhref{https://github.com/EleutherAI/lm-evaluation-harness}{\texttt{lm-evaluation-harness}} framework was utilized with the following configurations:

\begin{itemize}
    \item MMLU-Pro: Evaluated using the \myhref{https://github.com/EleutherAI/lm-evaluation-harness/tree/main/lm_eval/tasks/mmlu_pro}{\texttt{mmlu\_pro}} task configuration with consistent prompt design;
    \item BBH: Evaluated using the \myhref{https://github.com/EleutherAI/lm-evaluation-harness/tree/main/lm_eval/tasks/bbh}{\texttt{bbh}} task configuration, with chain-of-thought prompting given its effectiveness for multi-step reasoning tasks, and few-shot evaluation;
    \item MATH: Evaluated using the \myhref{https://github.com/EleutherAI/lm-evaluation-harness/tree/main/lm_eval/tasks/minerva_math}{\texttt{minerva\_math}} task configuration, with 4-shot evaluation.
\end{itemize}

We used \myhref{https://github.com/huggingface/accelerate}{\texttt{accelerate}} for efficient parallel processing while ensuring reproducibility through consistent evaluation procedures.

\subsection{Results and Analysis}
\subsubsection{Single-Method Compression Performance}
\begin{figure}[h]
\centering
\subfigure[LLaMA-3.1-8B]{
    \includegraphics[width=0.465690\columnwidth]{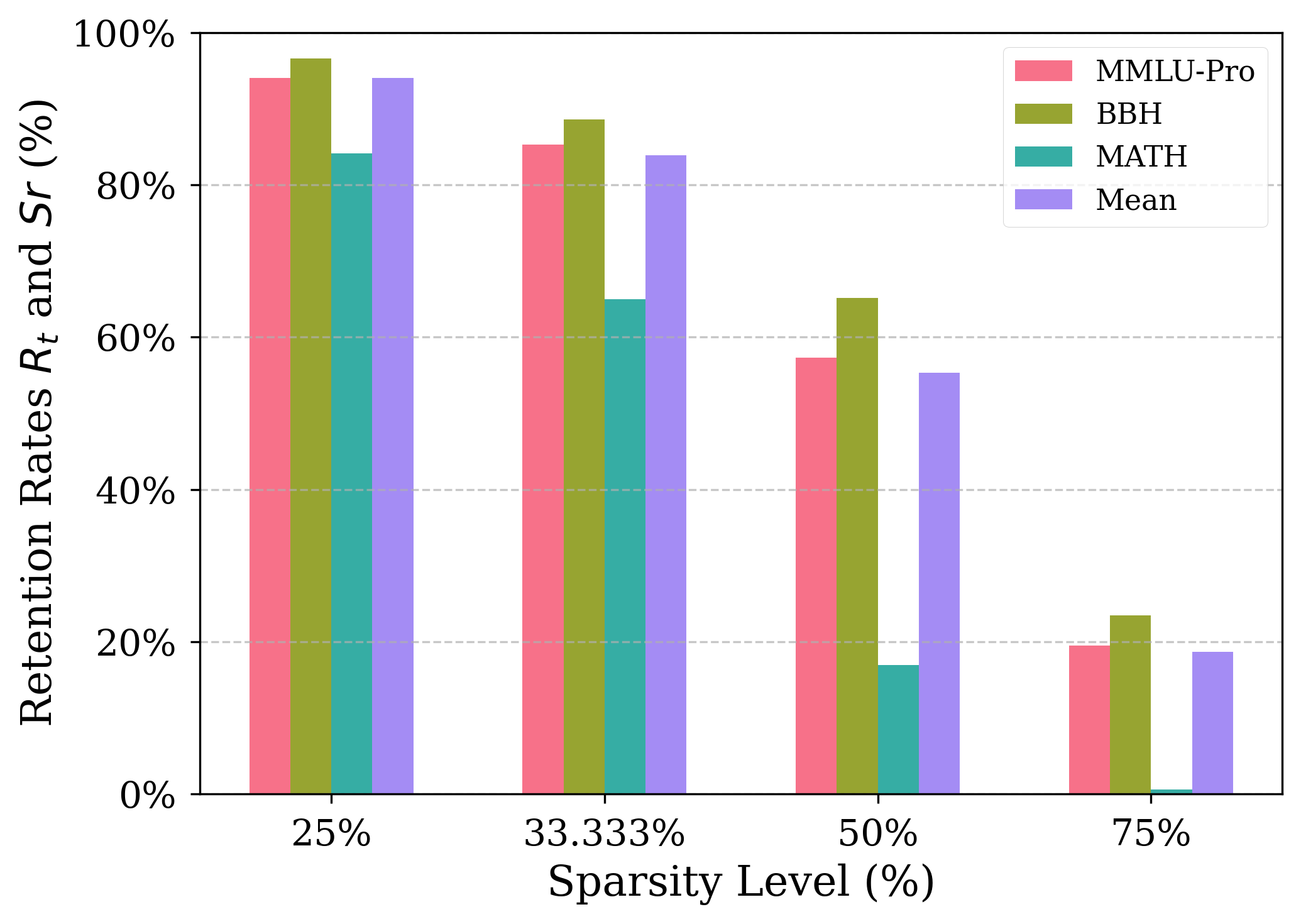}
    \label{fig:pruning_llama}
}
\subfigure[Mistral-7B-v0.3]{
    \includegraphics[width=0.465690\columnwidth]{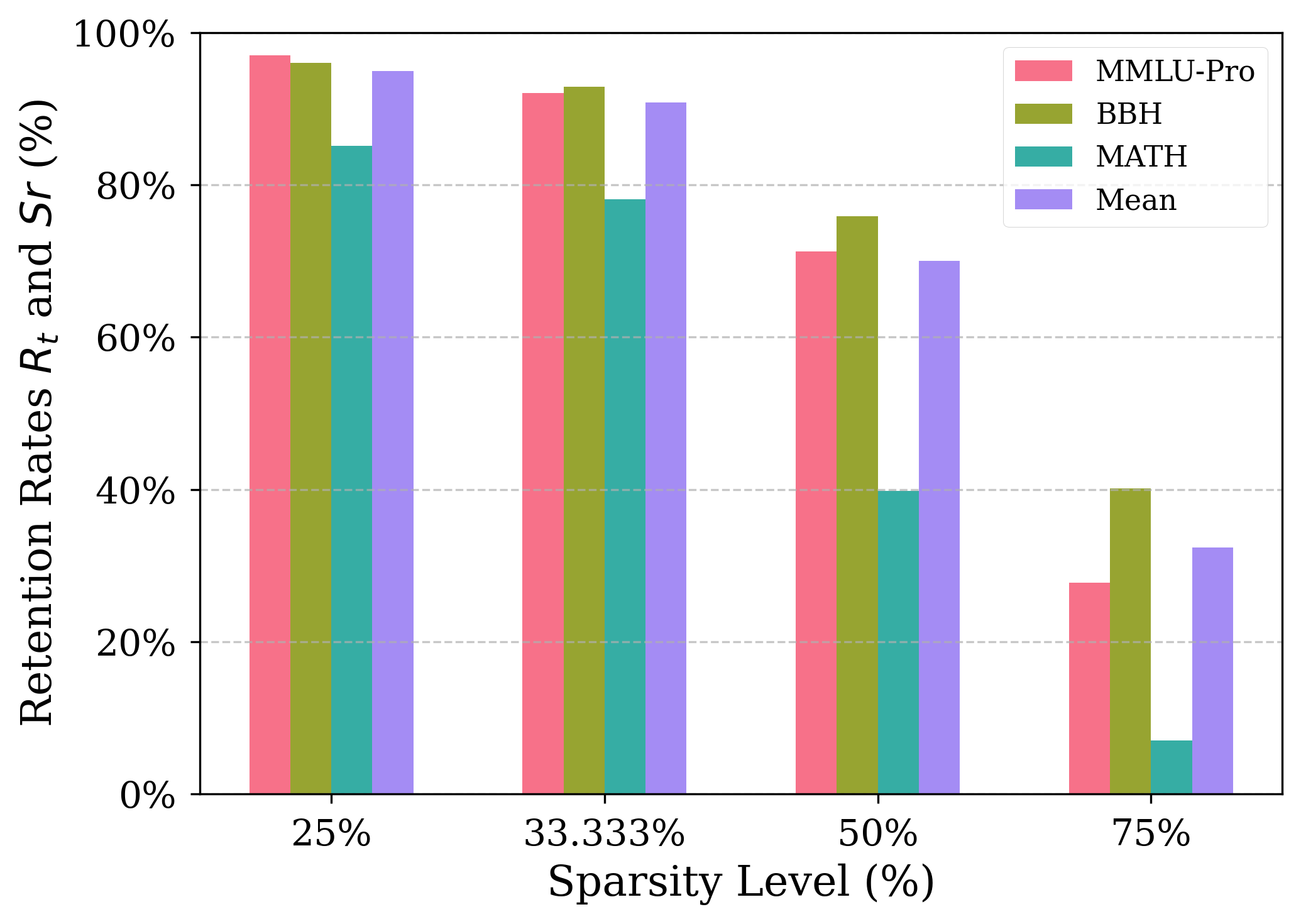}
    \label{fig:pruning_mistral}
}
\caption{Retention rates ($R_t$ across tasks and $Sr$) with pruning-only at different sparsity levels (25\%, 33.333\%, 50\%, and 75\%).}
\label{fig:pruning}
\end{figure}

Our experimental results demonstrate distinct patterns in compression behavior across different methods and models. For pruning (Figs.~\myref{fig:pruning_llama} and~\myref{fig:pruning_mistral}), both models exhibit graceful degradation up to 33.333\% sparsity.

\begin{figure}[h]
\centering
\subfigure[LLaMA-3.1-8B]{
    \includegraphics[width=0.465690\columnwidth]{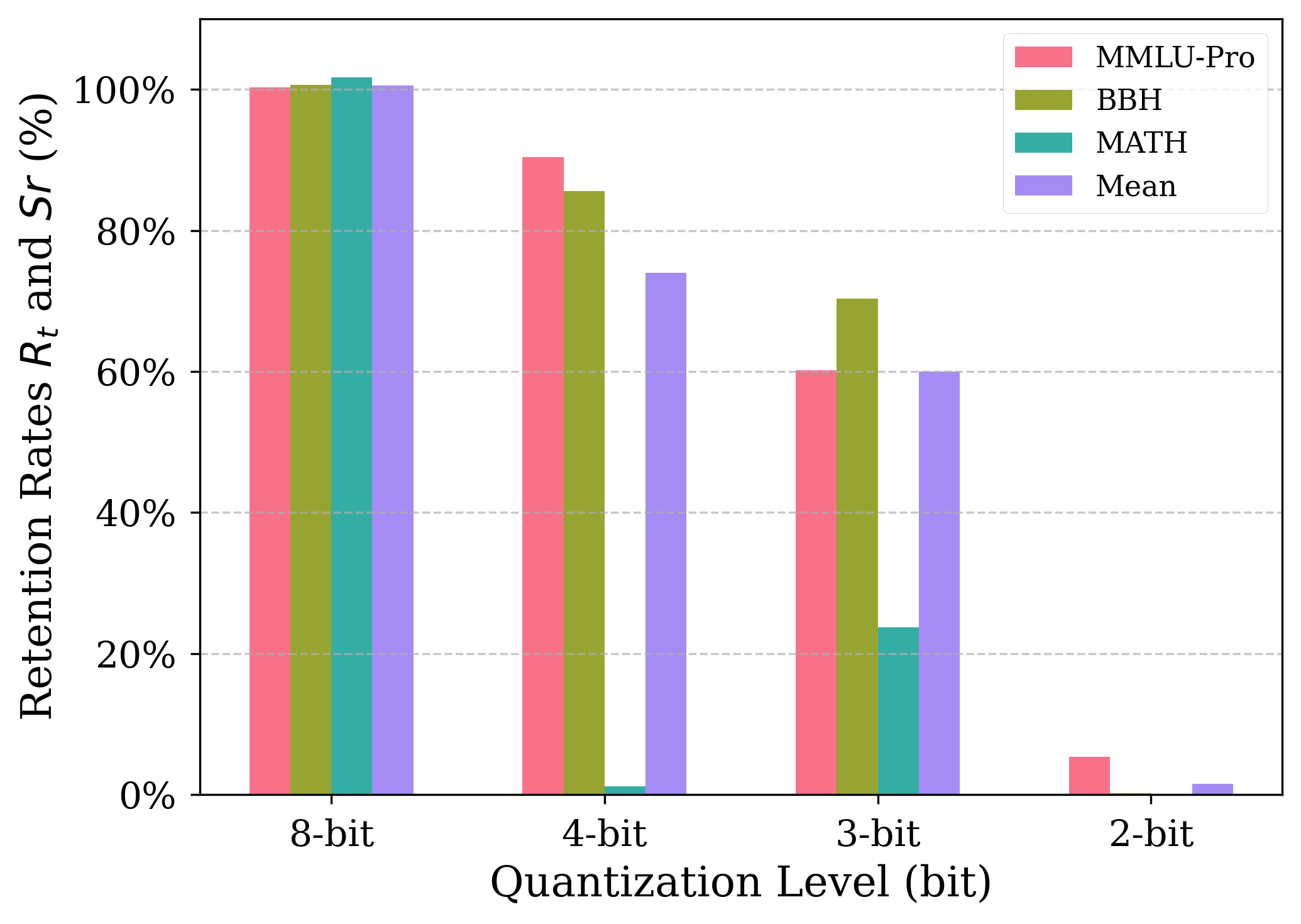}
    \label{fig:quant_llama}
}
\subfigure[Mistral-7B-v0.3]{
    \includegraphics[width=0.465690\columnwidth]{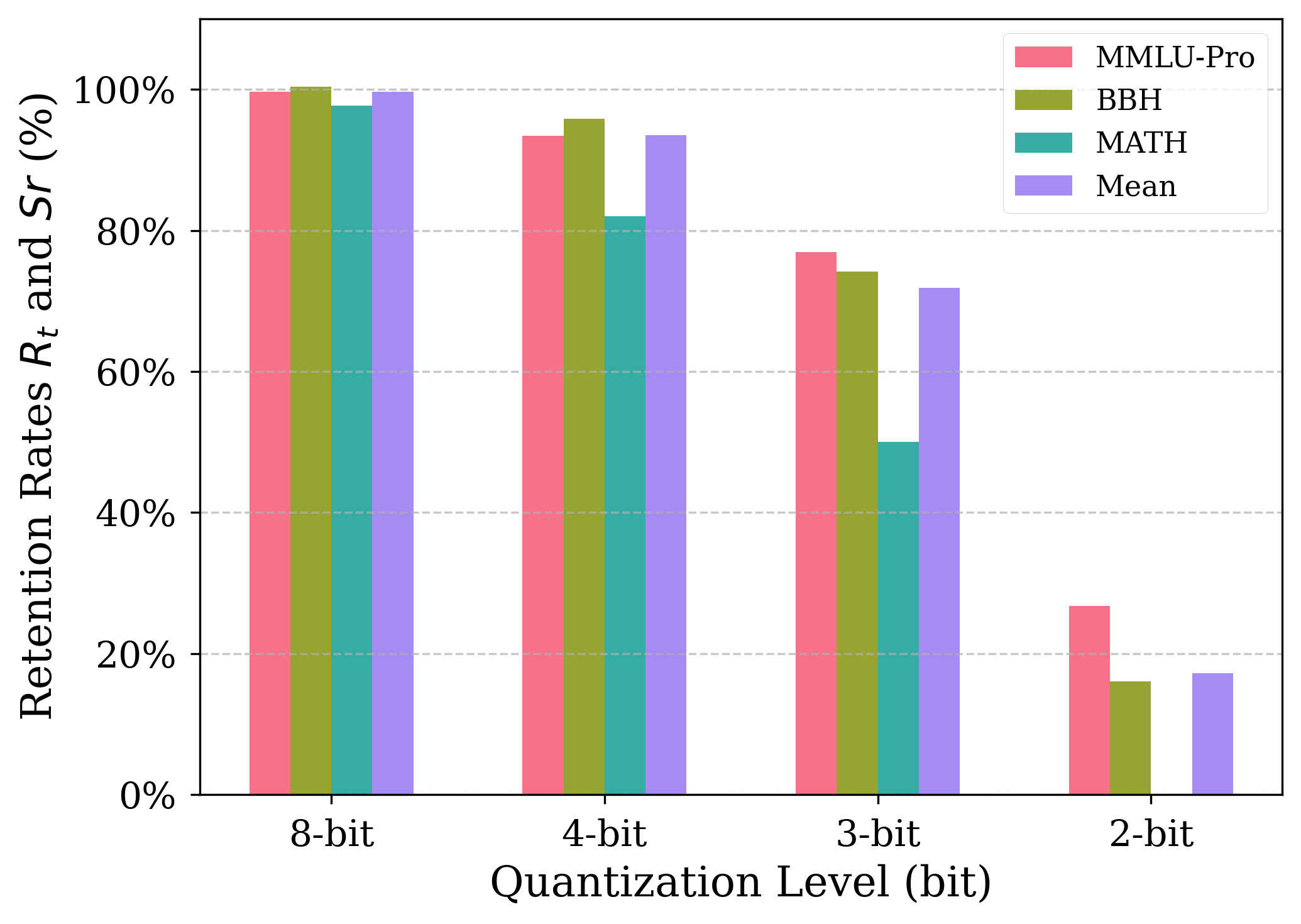}
    \label{fig:quant_mistral}
}
\caption{Retention rates ($R_t$ across tasks and $Sr$) with quantization-only at different bit-widths (8, 4, 3, and 2).}
\label{fig:quant}
\end{figure}

For quantization (Figs.~\myref{fig:quant_llama} and~\myref{fig:quant_mistral}), both models maintain near-baseline performance at 8-bit precision. Performance degradation becomes apparent at lower bit-widths, with a critical threshold around 3-bit precision where information loss begins to severely impact model capabilities.

Quantization consistently achieves greater performance than pruning at the same $TCr$ values. For instance, 8-bit quantization outperforms 50\% pruning despite both achieving the same 50\% $TCr$, while 4-bit quantization maintains better performance than 75\% pruning at 75\% $TCr$. This observation motivates our use of quantization as the baseline for evaluating joint compression strategies.

\begin{figure}[h]
\centering
\subfigure[Pruning-Only -- $SrCr_p$]{
    \includegraphics[width=0.465690\columnwidth]{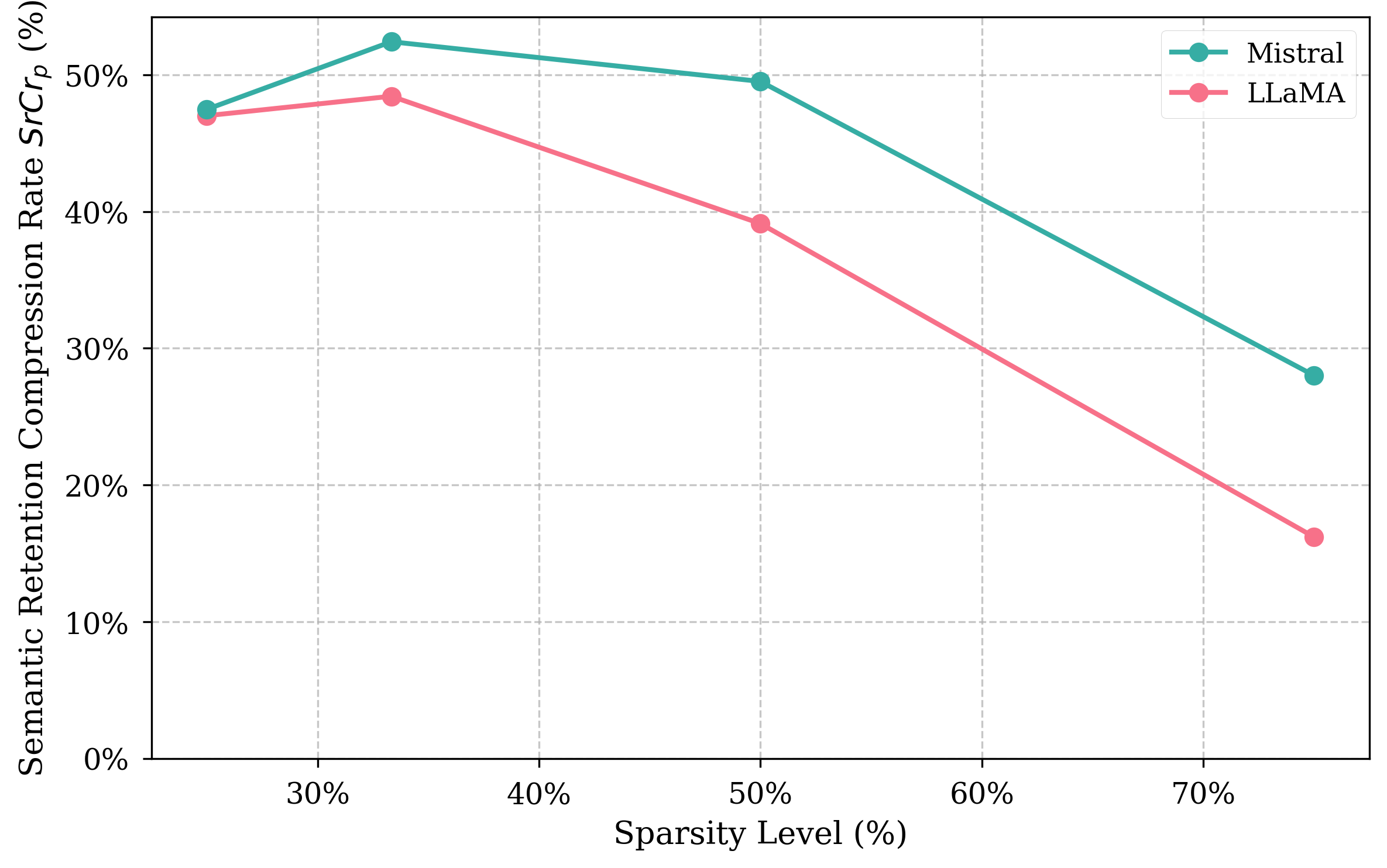}
    \label{fig:SrCr_p}
}
\subfigure[Quantization-Only -- $SrCr_q$]{
    \includegraphics[width=0.465690\columnwidth]{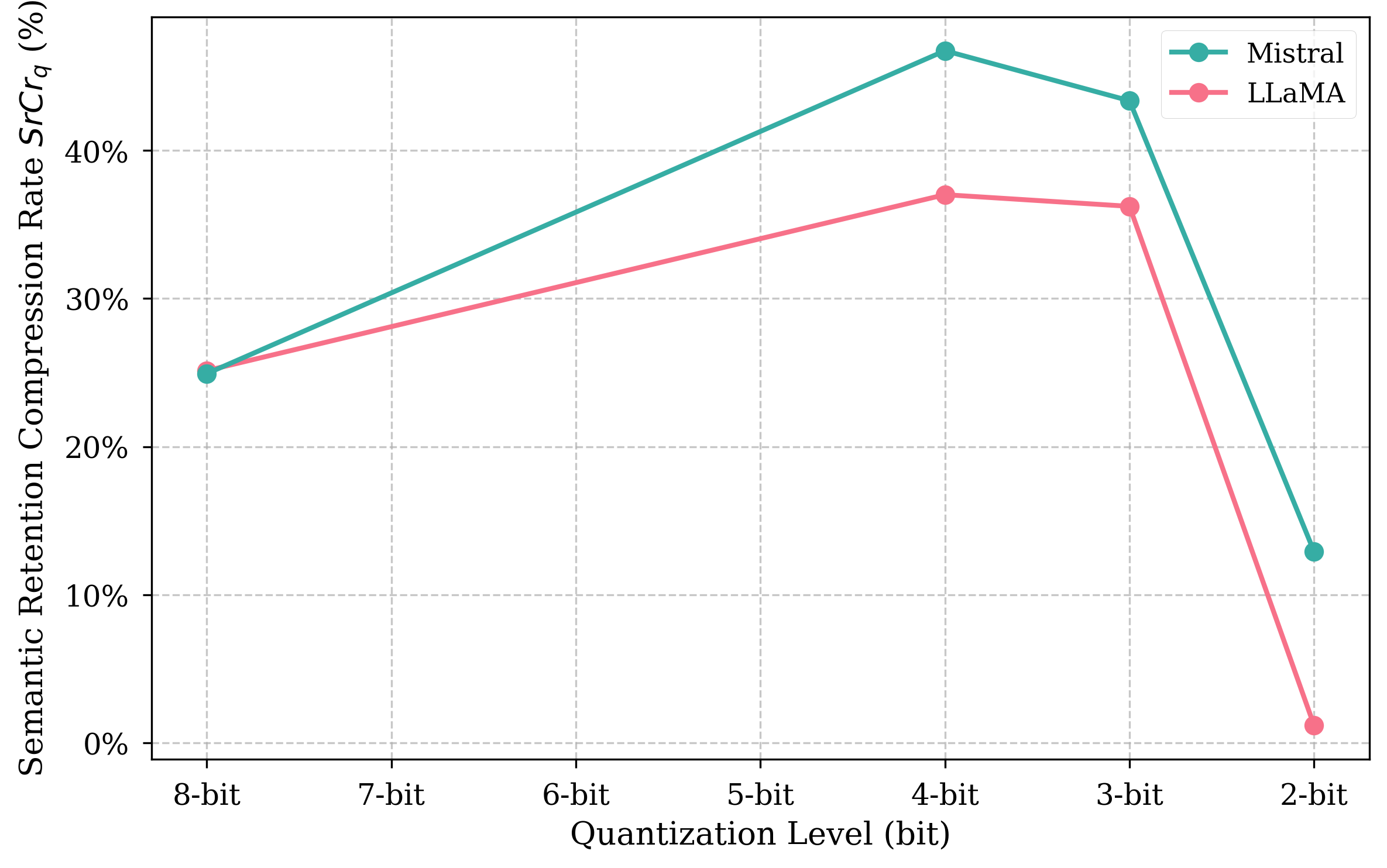}
    \label{fig:SrCr_q}
}
\caption{Semantic retention compression rates for single compression methods.}
\label{fig:single_srcr}
\end{figure}

Based on empirical observations of retention rates across setups, we developed the $SrCr$ metrics (Figs.~\myref{fig:SrCr_p} and~\myref{fig:SrCr_q}) to capture the balance between compression and performance retention in \myeqref{eq:srcr_p}\textcolor{red}{--}\myeqref{eq:srcr_j}. These metrics indicate that relatively low sparsity levels (25--33.333\%) and moderate quantization levels (4--3-bit) tend to maintain better compression-performance trade-offs compared to other approaches.

\medskip
\subsubsection{Validation of Error Approximation}
\begin{figure}[h]
\centering
\subfigure[LLaMA-3.1-8B]{
    \includegraphics[width=0.952159\columnwidth]{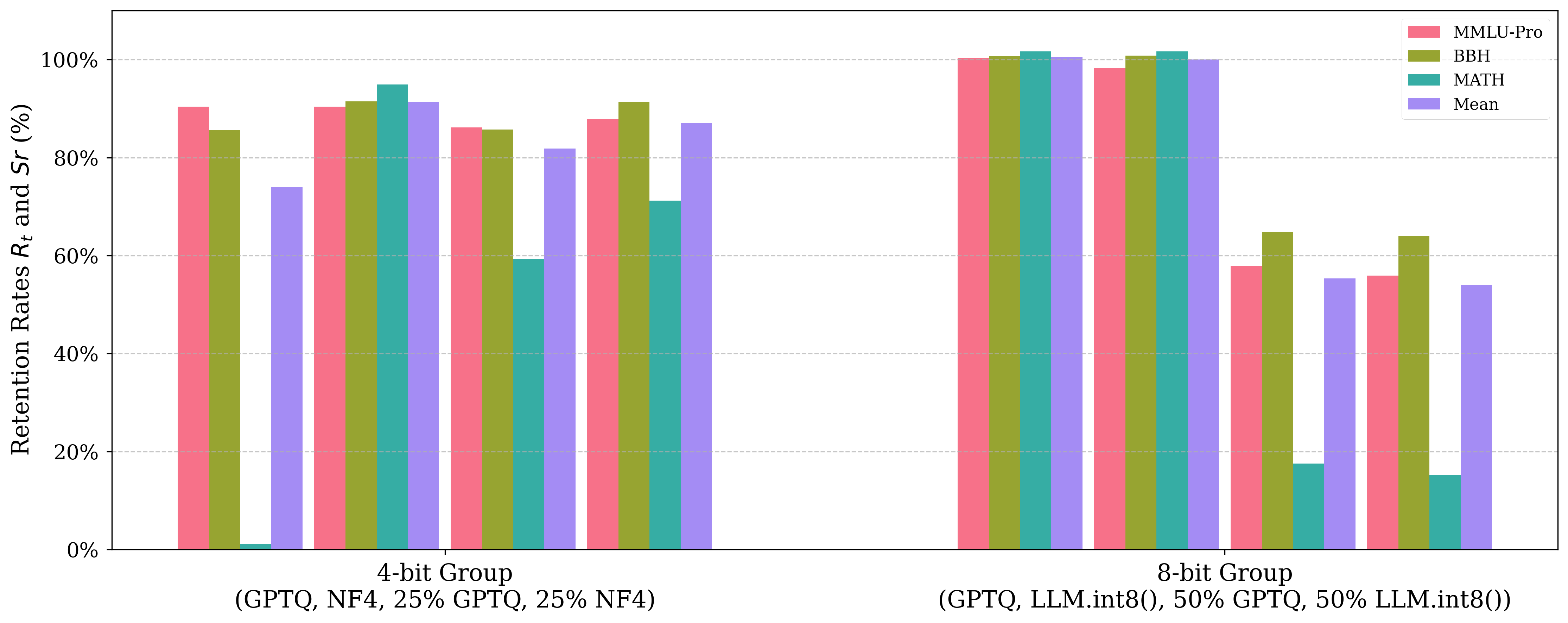}
    \label{fig:error_llama}
}
\subfigure[Mistral-7B-v0.3]{
    \includegraphics[width=0.952159\columnwidth]{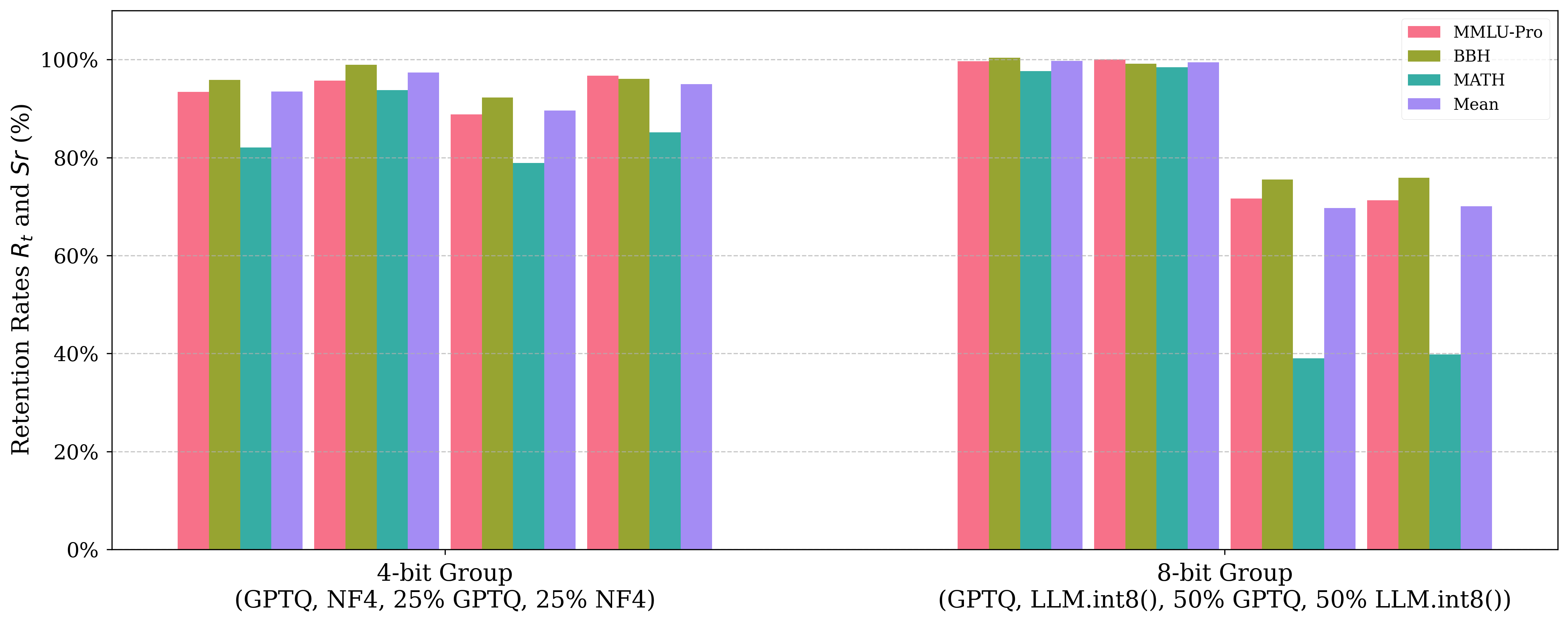}
    \label{fig:error_mistral}
}
\caption{Error $\|\delta_{A,j}\|_2^2$ analysis by comparison with NF4 and LLM.int8().}
\label{fig:error}
\end{figure}

Figs.~\myref{fig:error_llama} and~\myref{fig:error_mistral} compare GPTQ with simpler quantization techniques (NF4 and LLM.int8()) before and after pruning, revealing insights supporting our theoretical analysis. The similar relative degradation between GPTQ and simpler techniques across scenarios corroborates our error term decomposition in \myeqref{eq:error_a}\textcolor{red}{--}\myeqref{eq:error_a_simple}, while the consistent performance patterns between techniques suggest that fundamental information loss from quantization dominates technique-specific error characteristics. The relatively small performance differences between GPTQ and NF4/LLM.int8() suggest the accumulated error term $\|\delta_{A,j}\|_2^2$ remains well-bounded across configurations. This corroborates our use of Case~A sequential compression as a practical approximation for studying joint compression effects, while acknowledging that true joint optimization techniques may yield better performance through more sophisticated error distribution strategies.

\medskip
\subsubsection{Joint Compression Analysis}
We focus on three critical compression points highlighted in Table~\myref{tab:tcr_table}: 75\% $TCr$, 81.25\% $TCr$, and 87.5\% $TCr$. These represent key thresholds at which compression-performance trade-offs are most notable.

\begin{table}[h]
\caption{Theoretical Compression Rate Table}
\label{tab:tcr_table}
\begin{center}
\setlength{\tabcolsep}{4pt} % Reduce column separation
\begin{tabular}{|c|c|c|c|c|}
\hline
\diagbox[width=8em]{\textbf{Quant.}}{\textbf{Pruning}} & \textbf{0\%} & \textbf{25\%} & \textbf{33.333\%} & \textbf{50\%} \\
\hline
16 bit & 0\% & 25\% & 33.333\% & 50\% \\
8 bit & 50\% & 62.5\% & 66.667\% & \cellcolor{matchRed}{75\%} \\
4 bit & \cellcolor{matchRed}{75\%} & \cellcolor{matchBlue}{81.25\%} & 83.333\% & 87.5\% \\
3 bit & \cellcolor{matchBlue}{81.25\%} & 85.9375\% & \cellcolor{matchGreen}{87.5\%} & 90.625\% \\
2 bit & \cellcolor{matchGreen}{87.5\%} & 90.625\% & 91.667\% & 93.75\% \\
\hline
\end{tabular}
\end{center}
\end{table}

\begin{figure}[h]
\centering
\subfigure[LLaMA-3.1-8B]{
    \includegraphics[width=0.904991\columnwidth]{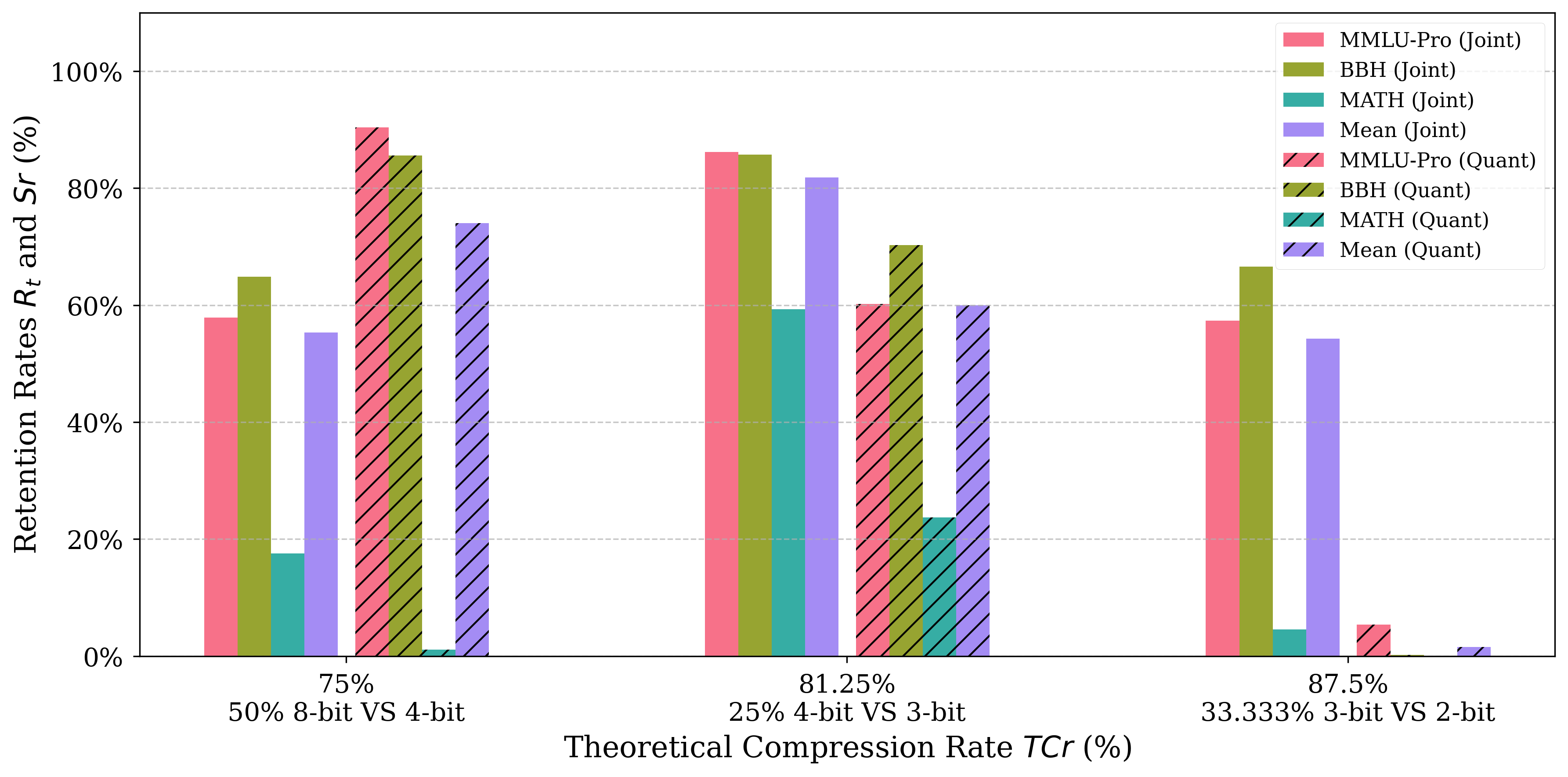}
    \label{fig:joint_llama}
}
\subfigure[Mistral-7B-v0.3]{
    \includegraphics[width=0.904991\columnwidth]{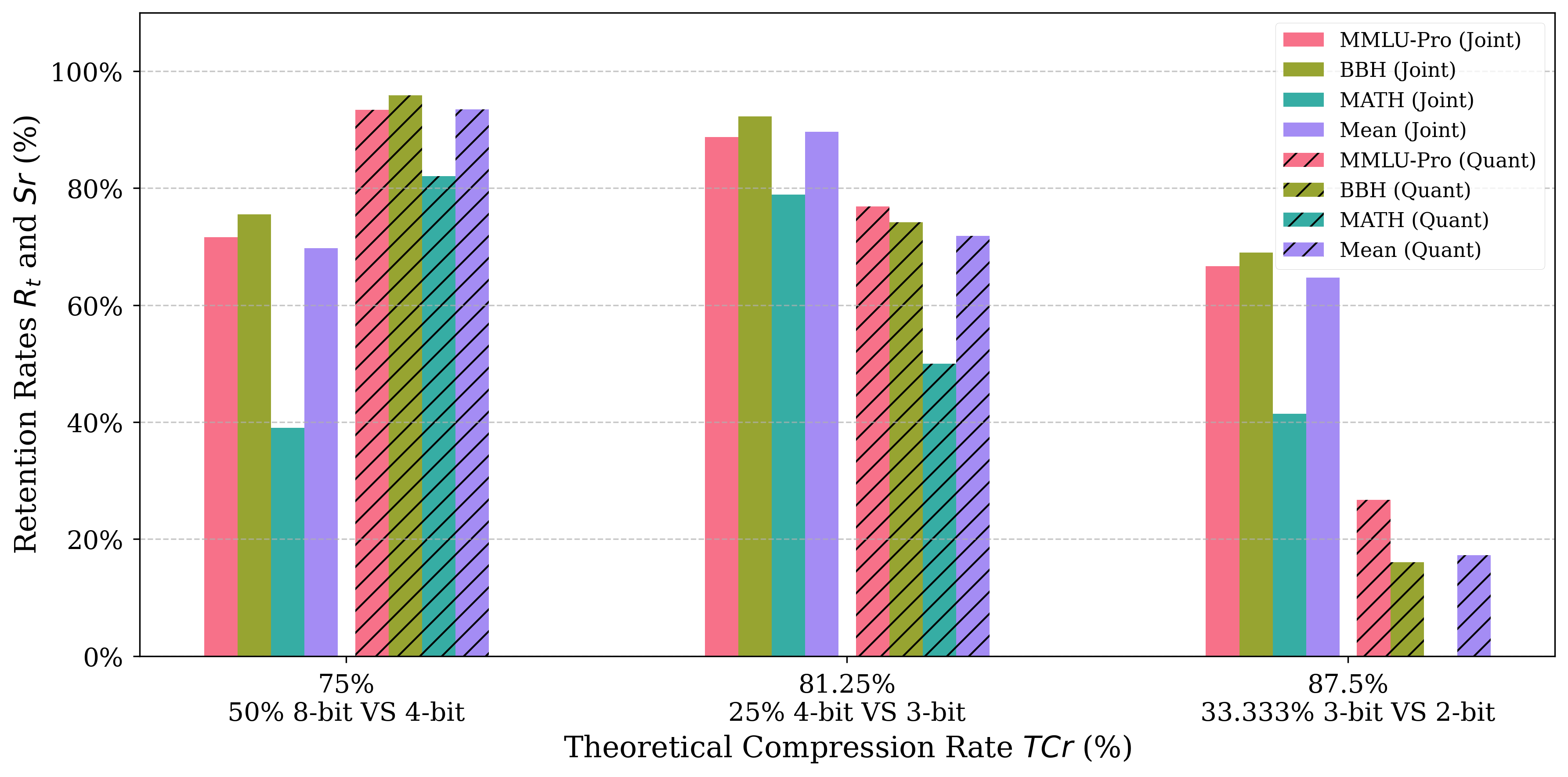}
    \label{fig:joint_mistral}
}
\caption{Retention rates ($R_t$ across tasks and $Sr$) for joint compression vs. quantization-only at
different $TCr$.}
\label{fig:joint}
\end{figure}

Figs.~\myref{fig:joint_llama} and~\myref{fig:joint_mistral} reveal several key insights:
\begin{enumerate}
\item At 75\% $TCr$, combining 50\% pruning with 8-bit quantization underperforms pure 4-bit quantization. This suggests that aggressive pruning (50\%) combined with conservative quantization (8-bit) results in an imbalanced configuration where the high sparsity level dominates the compression's impact on model performance.
\item At 81.25\% $TCr$, combining 25\% pruning with 4-bit quantization significantly outperforms pure 3-bit quantization, achieving around 20\% higher semantic retention across all evaluation metrics. This configuration offers a more balanced pruning-quantization trade-off.
\item At 87.5\% $TCr$, while 33.333\% pruning with 3-bit quantization shows degradation, it still substantially outperforms pure 2-bit quantization, maintaining meaningful capabilities where standalone methods fail entirely.
\end{enumerate}

\begin{figure}[h]
\centerline{\includegraphics[width=1\columnwidth]{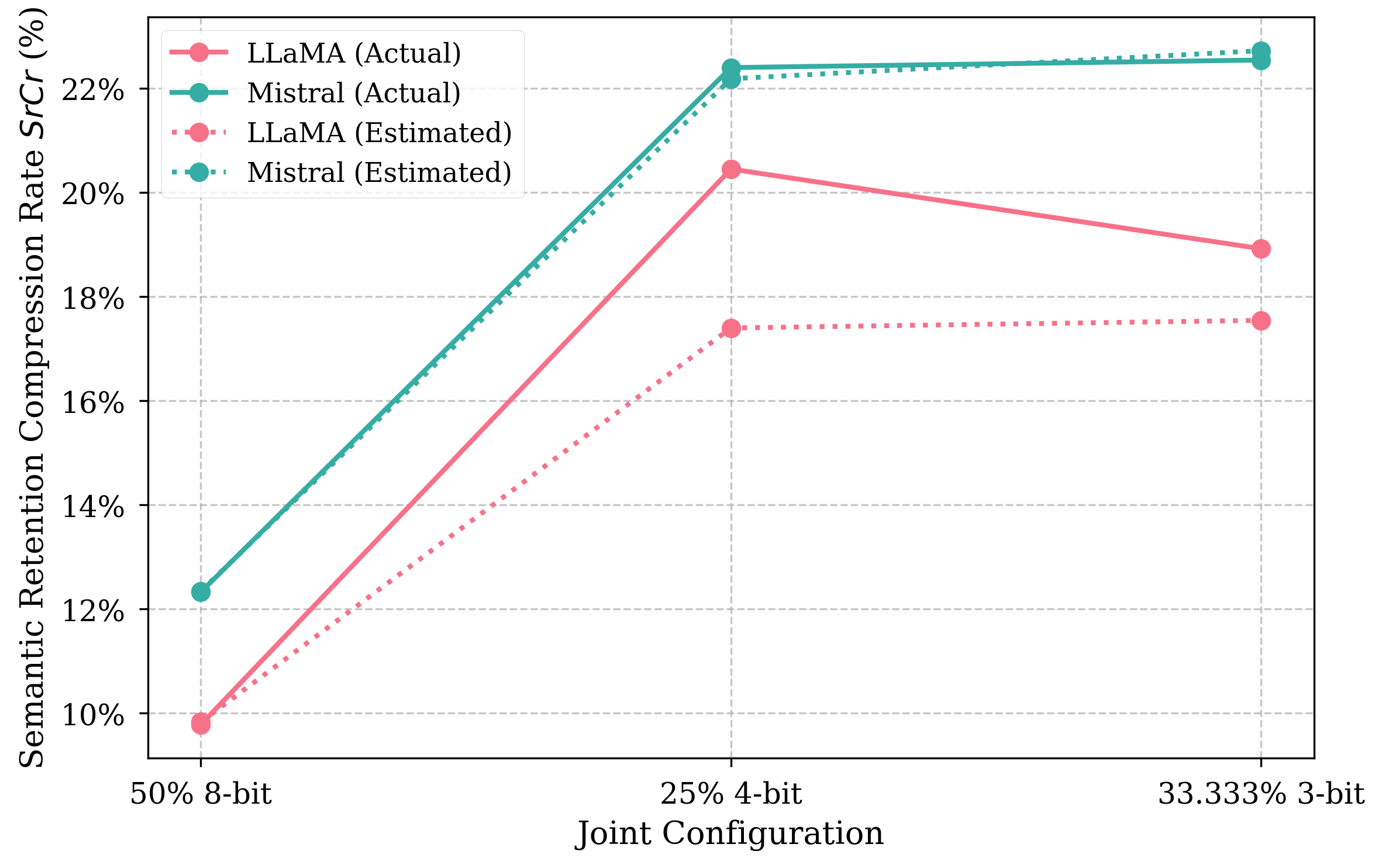}}
\caption{Semantic retention compression rate in joint compression ($SrCr_j$) and its estimate ($SrCr_p * SrCr_q$).}
\label{fig:joint_srcr_vs}
\end{figure}

The $SrCr_j$ metric (Fig.~\myref{fig:joint_srcr_vs}) shows that combining 25\% pruning with 4-bit quantization provides the optimal compression-performance trade-off for both models. Additionally, Mistral maintains strong performance even with the more aggressive 33.333\% pruning and 3-bit quantization configuration, while LLaMA shows significant degradation under these settings.

The comparison between actual and estimated joint compression effectiveness reveals that Mistral's estimated performance closely matches its actual performance, while LLaMA's actual performance exceeds its estimated performance. This is largely due to LLaMA’s unexpected MATH performance drop under 4-bit quantization, which skews the estimated $SrCr$.

\medskip
\subsubsection{Semi-Structured Pruning Investigation}
\paragraph{Motivation and Hardware Considerations}
Recent GPU architectures, particularly NVIDIA's Ampere platform, demonstrate native support for semi-structured sparsity patterns, notably 2:4 sparsity~\cite{sparsegpt}. While some patterns investigated in our study (1:4 and 2:8) are not currently accelerable on hardware, they are likely to be supported in the near future, unlike other patterns (1:3 and 2:6), which we include solely for comprehensiveness and comparison purposes.

Semi-structured patterns offer benefits beyond current hardware, as they:
\begin{itemize}
\item Reduce memory access and computational complexity;
\item Enable more efficient sparse matrix operations on GPU Tensor Cores;
\item Enable specialized hardware acceleration;
\item Complement quantization through improved memory bandwidth utilization;
\item Reduce memory overhead compared to unstructured pruning's full mask matrix requirement (1 bit per weight), with semi-structured pruning requiring only $\frac{\log_2\binom{m}{n}}{m}$ bits per weight in the most efficient implementation.
\end{itemize}

Based on our findings from unstructured pruning experiments, where 50\% pruning with 8-bit quantization significantly underperformed pure 4-bit quantization at 75\% TCr, we focused our semi-structured pruning investigation on the more promising 25\% and 33.333\% sparsity levels, excluding the 50\% configurations. Indeed, semi-structured pruning inherently imposes stricter constraints than unstructured pruning, typically resulting in lower performance at equivalent sparsity levels. Given these constraints, the already underperforming 50\%-8bit combination in the unstructured case offered no potential for viable performance when applied with semi-structured patterns.

\begin{figure}[h]
\centering
\subfigure[LLaMA-3.1-8B]{
    \includegraphics[width=0.797142\columnwidth]{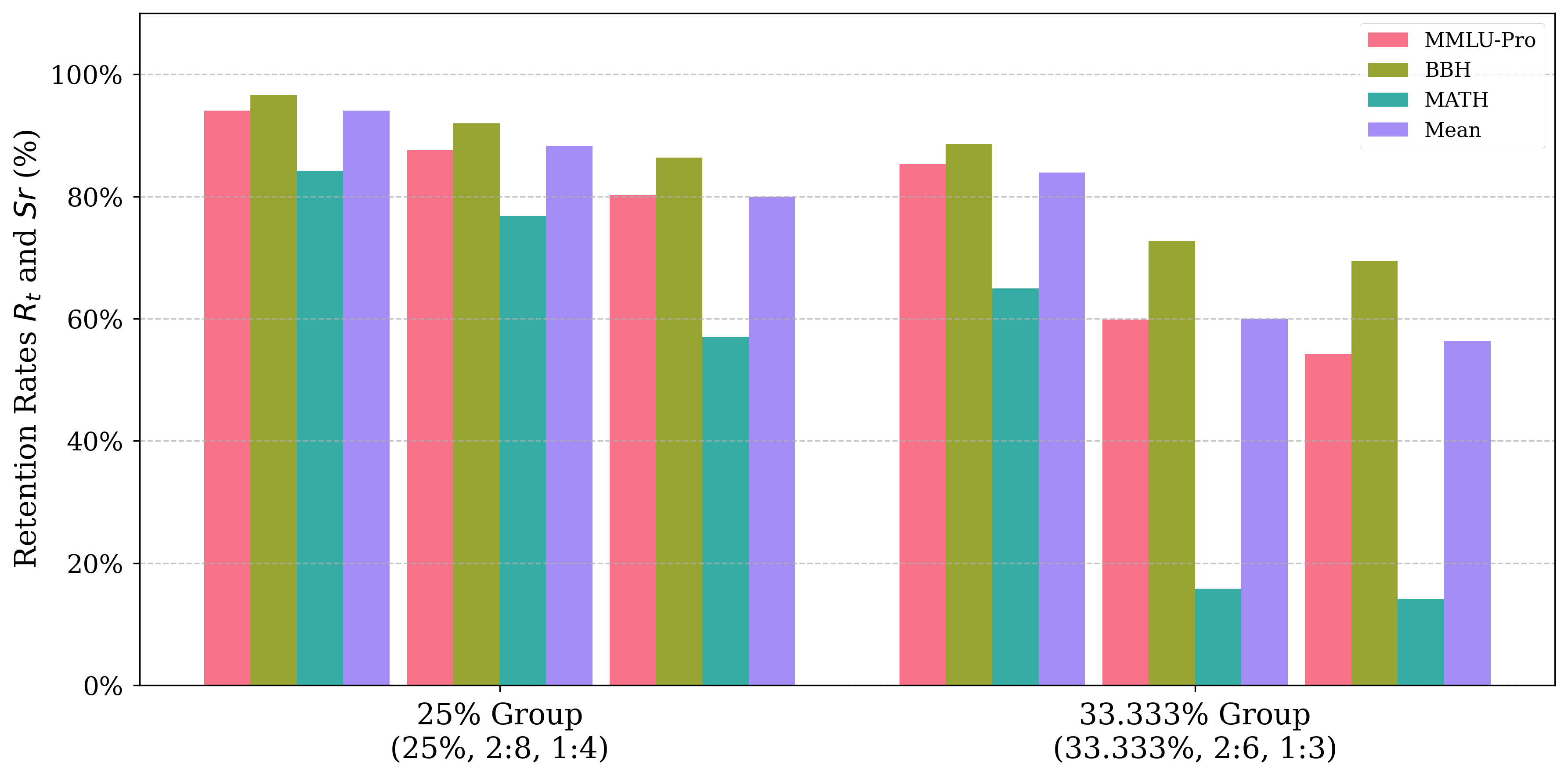}
    \label{fig:struct_llama}
}
\subfigure[Mistral-7B-v0.3]{
    \includegraphics[width=0.797142\columnwidth]{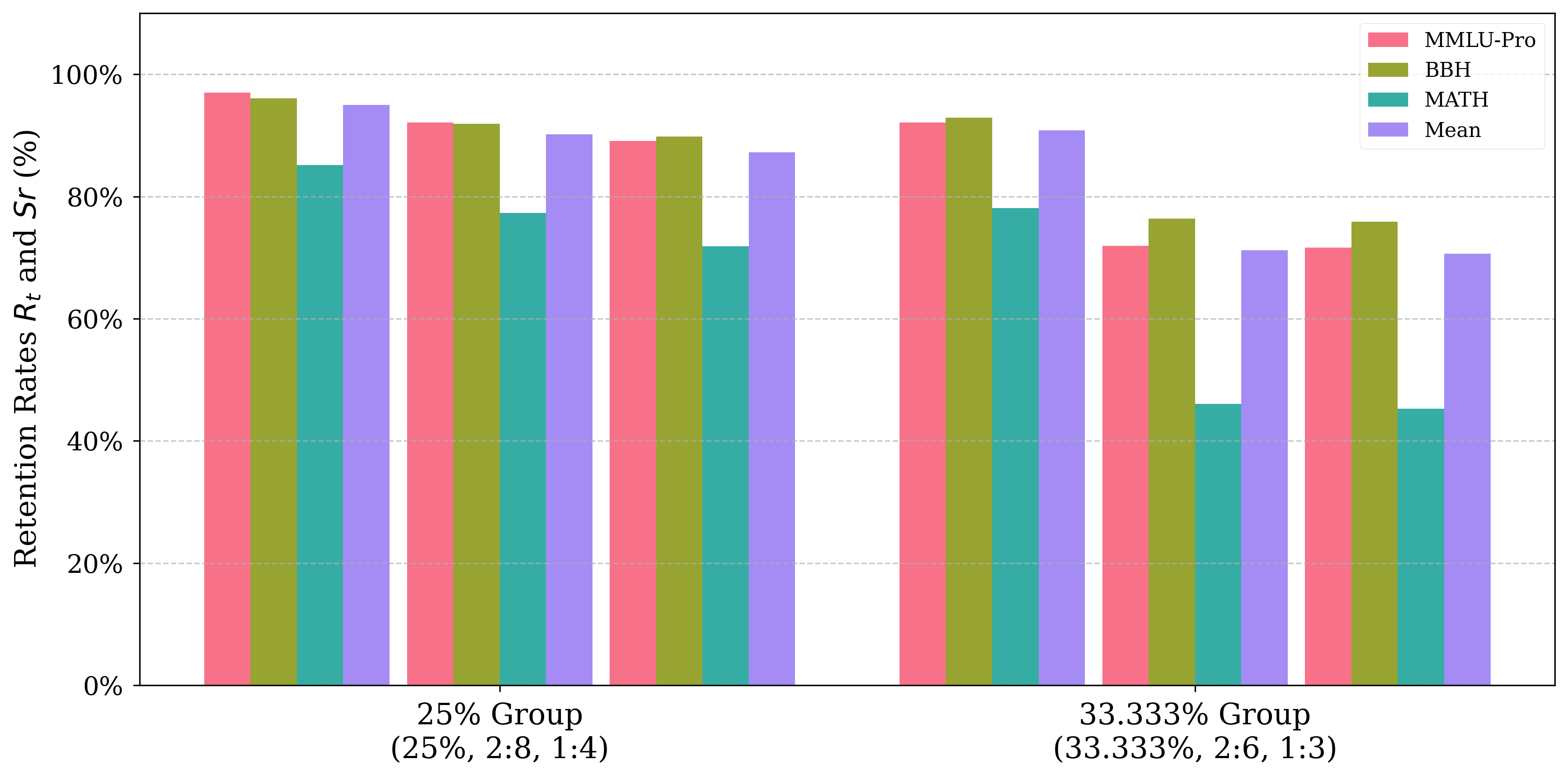}
    \label{fig:struct_mistral}
}
\caption{Retention rates ($R_t$ across tasks and $Sr$) for pruning-only with and without semi-structured patterns.}
\label{fig:struct_pruning}
\end{figure}

\smallskip
\paragraph{Performance Analysis}
Our experiments with semi-structured pruning patterns (Figs.~\myref{fig:struct_llama} and~\myref{fig:struct_mistral}) show that at 25\% sparsity, all patterns maintain relatively strong performance, with the 2:8 pattern emerging as particularly promising despite lacking current hardware acceleration. The impact of structured constraints becomes more pronounced at 33.333\% sparsity, especially with more rigid patterns like 2:6 and 1:3.

\begin{figure}[h]
\centering
\subfigure[LLaMA-3.1-8B]{
    \includegraphics[width=0.797142\columnwidth]{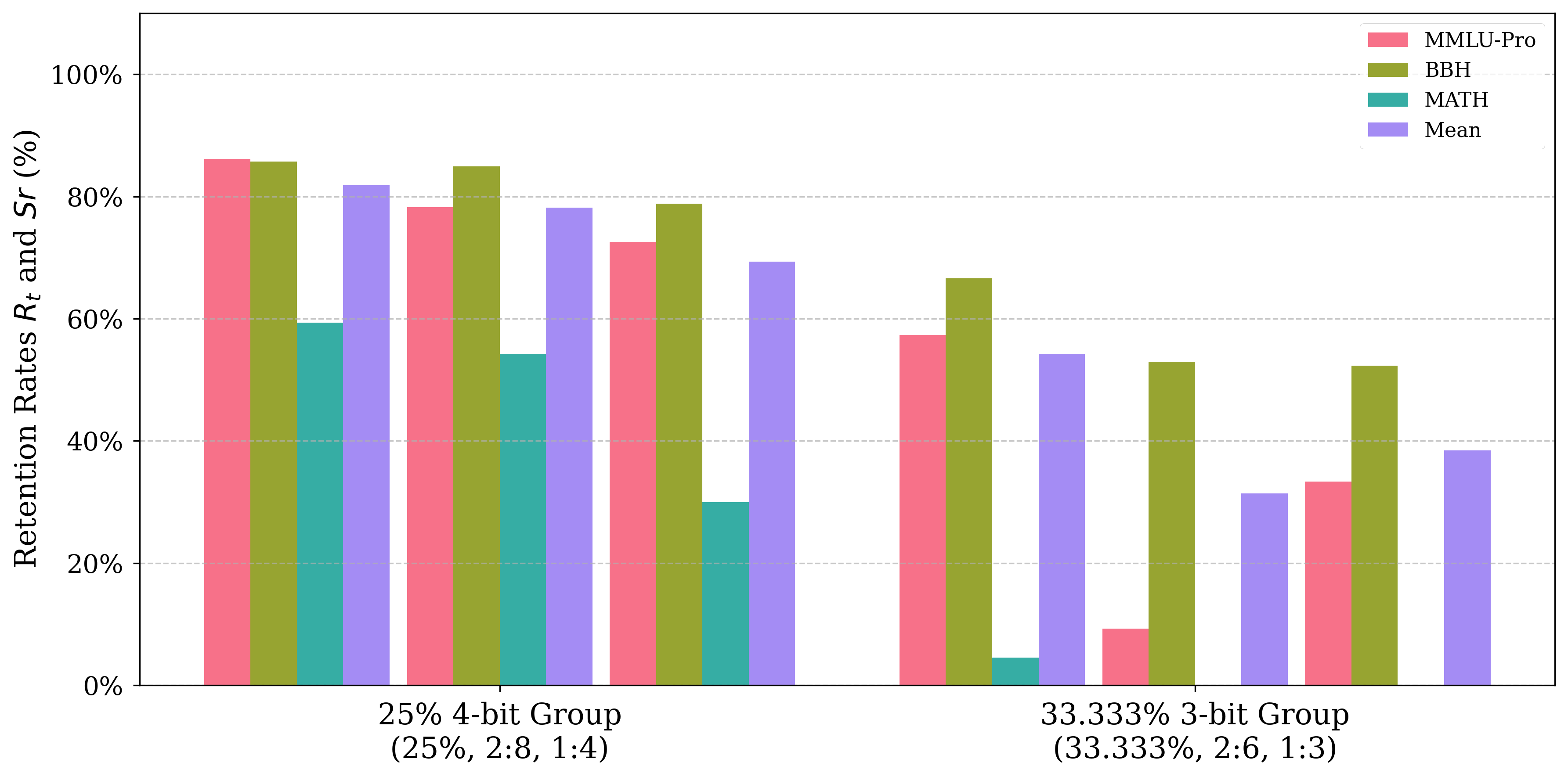}
    \label{fig:struct_joint_llama}
}
\subfigure[Mistral-7B-v0.3]{
    \includegraphics[width=0.797142\columnwidth]{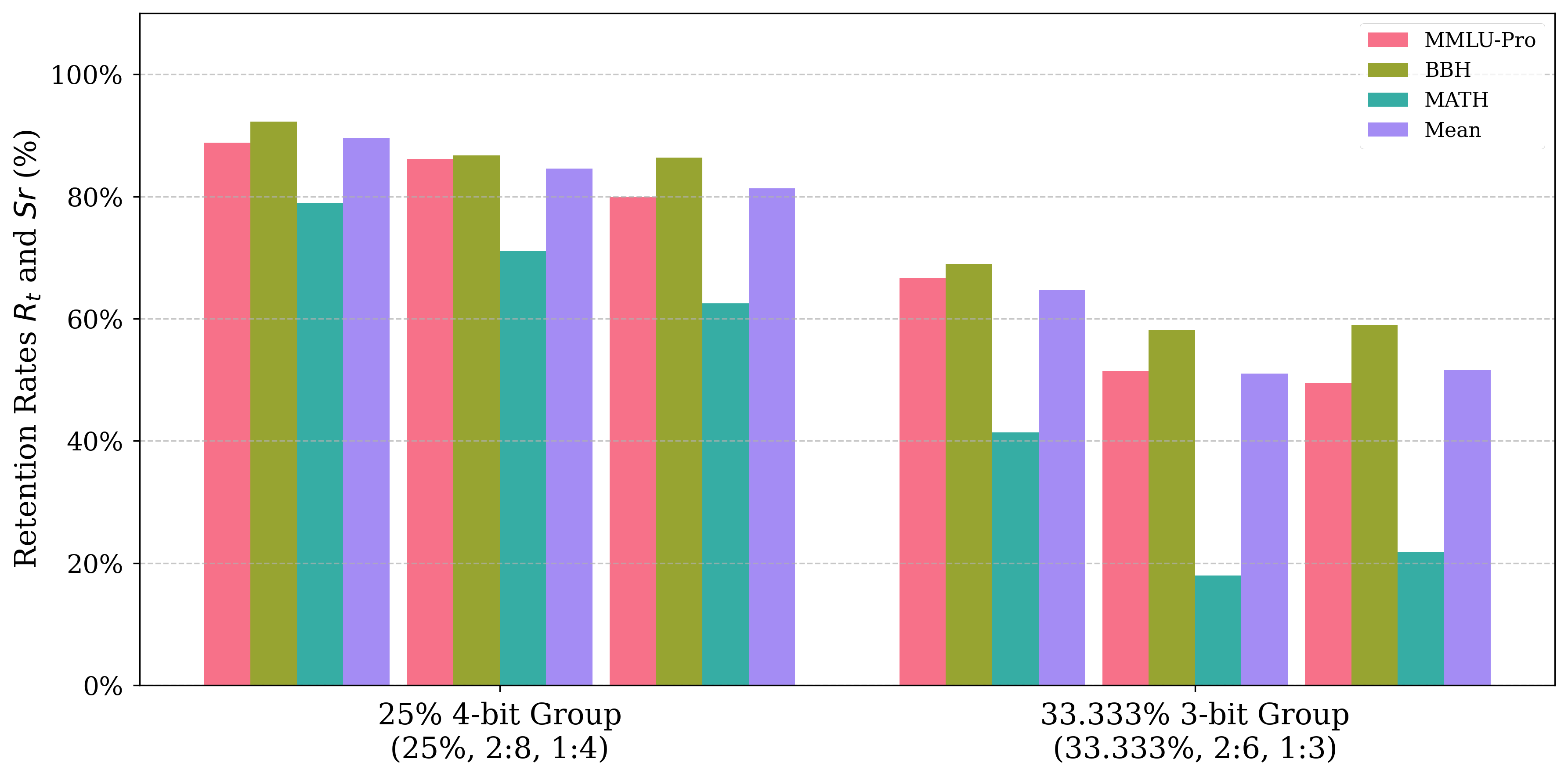}
    \label{fig:struct_joint_mistral}
}
\caption{Retention rates ($R_t$ across tasks and $Sr$) for joint compression with and without semi-structured patterns.}
\label{fig:struct_joint}
\end{figure}

In joint compression scenarios (Figs.~\myref{fig:struct_joint_llama} and~\myref{fig:struct_joint_mistral}), semi-structured patterns maintain strong performance when combined with moderate quantization (25\% with 4-bit), significantly outperforming pure 3-bit quantization at the same $TCr$ value. This trend persists even at higher compression rates, where constrained patterns with 3-bit quantization still outperform pure 2-bit quantization.

\medskip
\subsubsection{Key Observations}
Our comprehensive analysis reveals several fundamental patterns across compression methods:

\begin{itemize}
    \item Model Architecture Impact: Mistral consistently demonstrates greater resilience to compression compared to LLaMA---particularly under aggressive configurations. This difference may be attributed to two key factors:
    \begin{itemize}
        \item Architecture and Size: LLaMA's slightly larger parameter count (8B vs. 7B) may result in different information density characteristics;
        \item Multilingual vs. Monolingual: LLaMA's multilingual capabilities likely require more complex internal representations, making accurate compression harder compared to Mistral's English-focused architecture.
    \end{itemize}
    
    \item Task Sensitivity: Mathematical reasoning, instruction following, and output-format adherence appear to rely more heavily on the full parameter space compared to other cognitive tasks, as shown by MATH's consistently higher sensitivity to compression across all experiments.
\end{itemize}

\medskip
\subsubsection{Practical Implications}
Our analysis yields several key implications for practical deployment:

\begin{itemize}
    \item Joint compression approaches, particularly balanced configurations like 25\% pruning with 4-bit quantization, offer superior compression-performance trade-offs compared to aggressive single-method compression;
    
    \item Semi-structured pruning patterns, especially 2:8, present a promising direction for hardware-efficient deployment while maintaining competitive performance when combined with moderate quantization;
    
    \item Model architecture significantly influences compression potential, with specialized models like Mistral demonstrating greater resilience to aggressive compression compared to more general-purpose models like LLaMA.
\end{itemize}

These results suggest that effective model compression requires careful consideration of both the compression configuration and the underlying model architecture, with joint approaches offering promising pathways for efficient deployment while maintaining model capabilities.

\section{Conclusion} \label{conclusion}
This work presents a systematic investigation of joint compression for LLMs, introducing a theoretical framework and novel metrics for evaluating compression effectiveness. Our experiments reveal that strategic combinations of pruning and quantization can achieve greater performance than single-method approaches at equivalent compression rates. Specifically, we demonstrate that moderate joint compression (25\% pruning with 4-bit quantization) consistently outperforms pure 3-bit quantization by approximately 20\% in semantic retention at the same Theoretical Compression Rate ($TCr$).

The introduction of Semantic Retention Compression Rate ($SrCr$) provides a principled approach to comparing different compression configurations, enabling rigorous evaluation of size-capability trade-offs. Our analysis of semi-structured pruning patterns reveals promising directions for hardware-efficient deployment, with patterns like 2:8 showing competitive performance when combined with quantization while potentially enabling better hardware utilization.

These findings suggest that future development of LLM compression techniques should focus on joint optimization approaches rather than pushing single methods to their limits. The demonstrated success of well-balanced compression configurations indicates a promising direction for making LLM deployment more practical in resource-constrained environments while preserving model capabilities.

\section{Limitations and Future Work} \label{limitations}
While our work provides valuable insights into joint compression, several limitations and opportunities for future research remain:

\subsection{Limitations}
First, our sequential approximation of joint compression, while practically useful, may not capture the full potential of truly simultaneous optimization of pruning and quantization. Second, our theoretical framework assumes independence between compression methods, which may not fully reflect their complex interactions. Third, our experiments focused exclusively on 7--8B models, and as parameter efficiency varies with model size and improves over time, our $SrCr$ formula may require adjustments to generalize across evolving architectures of all model scales. Fourth, as compression techniques themselves evolve, our $SrCr$ formula may require further refinement. Fifth, our evaluation metrics, while comprehensive, may not capture all aspects of model performance, particularly for specialized tasks or domain-specific applications.

\subsection{Future Work}
Several promising directions emerge from our findings:

(1) Extension of our theoretical framework to include other compression methods such as low-rank matrix factorization and knowledge distillation.

(2) Development of unified joint compression algorithms that simultaneously optimize pruning and quantization, potentially achieving better performance than sequential approaches. Such techniques would ideally incorporate other techniques' advances---like outlier handling, weight–activation redistribution, mixed-precision quantization, double quantization strategies, and structural pruning with minimal overhead.

(3) Investigation of hardware-aware optimization strategies that align compression with specific accelerator architectures.

(4) Exploration of dynamic compression techniques that adapt to computational requirements during inference.

These directions, combined with rapid advancements in LLM architectures, suggest a rich landscape for future research in efficient model deployment.

\section*{Acknowledgment}
The authors thank ESIEA for funding this research and providing access to resources, especially GPUs, as well as the resource management team for their support.

They also acknowledge the use of Claude 3.5 Sonnet (\mbox{Anthropic}, 2024) for assistance with academic writing refinement throughout the manuscript.

\clearpage
\onecolumn
\appendix

\begin{figure*}[h]
\centering
\makebox[\textwidth][c]{

\begin{minipage}[t]{0.48\textwidth}
\begin{table}[H]
\centering
\parbox[c][3.7\baselineskip][c]{\textwidth}{
\caption{SparseGPT, Unstructured Pruning-Only Results}
\label{tab:sparsegpt_unstruct_perf}
}
\resizebox{\textwidth}{!}{
\begin{tabular}{|c|c|cc|cc|cc|cc|}
\hline
\multirow{2}{*}{\textbf{Model}} & \multirow{2}{*}{\textbf{Sparsity}} & \multicolumn{2}{c|}{\textbf{MMLU-Pro}} & \multicolumn{2}{c|}{\textbf{BBH}} & \multicolumn{2}{c|}{\textbf{MATH}} & \multicolumn{2}{c|}{\textbf{Mean}} \\
\cline{3-10}
& & \textbf{\textit{Score}} & \textbf{\textit{Std}} & \textbf{\textit{Score}} & \textbf{\textit{Std}} & \textbf{\textit{Score}} & \textbf{\textit{Std}} & \textbf{\textit{Score}} & \textbf{\textit{Std}} \\
\hline
\multirow{5}{*}{LLaMA 8B}
& 0\% & 35.4 & 0.4 & 62.3 & 0.5 & 17.7 & 0.5 & 38.5 & 0.3 \\
& 25\% & 33.3 & 0.4 & 60.2 & 0.5 & 14.9 & 0.5 & 36.2 & 0.3 \\
& 33.333\% & 30.2 & 0.4 & 55.2 & 0.5 & 11.5 & 0.4 & 32.3 & 0.3 \\
& 50\% & 20.3 & 0.4 & 40.6 & 0.6 & 3.0 & 0.2 & 21.3 & 0.2 \\
& 75\% & 6.9 & 0.2 & 14.6 & 0.4 & 0.1 & 0.0 & 7.2 & 0.2 \\
\hline
\multirow{5}{*}{Mistral 7B}
& 0\% & 30.3 & 0.4 & 58.0 & 0.5 & 12.8 & 0.5 & 33.7 & 0.3 \\
& 25\% & 29.4 & 0.4 & 55.7 & 0.6 & 10.9 & 0.4 & 32.0 & 0.3 \\
& 33.333\% & 27.9 & 0.4 & 53.9 & 0.6 & 10.0 & 0.4 & 30.6 & 0.3 \\
& 50\% & 21.6 & 0.4 & 44.0 & 0.5 & 5.1 & 0.3 & 23.6 & 0.2 \\
& 75\% & 8.4 & 0.3 & 23.3 & 0.5 & 0.9 & 0.1 & 10.9 & 0.2 \\
\hline
\end{tabular}
}
\end{table}
\end{minipage}

\begin{minipage}[t]{0.48\textwidth}
\begin{table}[H]
\centering
\parbox[c][3.7\baselineskip][c]{\textwidth}{
\caption{GPTQ, Quantization-Only Results}
\label{tab:gptq_perf}
}
\resizebox{\textwidth}{!}{
\begin{tabular}{|c|c|cc|cc|cc|cc|}
\hline
\multirow{2}{*}{\textbf{Model}} & \multirow{2}{*}{\textbf{Bits}} & \multicolumn{2}{c|}{\textbf{MMLU-Pro}} & \multicolumn{2}{c|}{\textbf{BBH}} & \multicolumn{2}{c|}{\textbf{MATH}} & \multicolumn{2}{c|}{\textbf{Mean}} \\
\cline{3-10}
& & \textbf{\textit{Score}} & \textbf{\textit{Std}} & \textbf{\textit{Score}} & \textbf{\textit{Std}} & \textbf{\textit{Score}} & \textbf{\textit{Std}} & \textbf{\textit{Score}} & \textbf{\textit{Std}} \\
\hline
\multirow{5}{*}{LLaMA 8B}
& 16 & 35.4 & 0.4 & 62.3 & 0.5 & 17.7 & 0.5 & 38.5 & 0.3 \\
& 8 & 35.5 & 0.4 & 62.7 & 0.5 & 18.0 & 0.5 & 38.7 & 0.3 \\
& 4 & 32.0 & 0.4 & 53.3 & 0.5 & 0.2 & 0.1 & 28.5 & 0.2 \\
& 3 & 21.3 & 0.4 & 43.8 & 0.6 & 4.2 & 0.3 & 23.1 & 0.2 \\
& 2 & 1.9 & 0.1 & 0.1 & 0.0 & 0.0 & 0.0 & 0.6 & 0.0 \\
\hline
\multirow{5}{*}{Mistral 7B}
& 16 & 30.3 & 0.4 & 58.0 & 0.5 & 12.8 & 0.5 & 33.7 & 0.3 \\
& 8 & 30.2 & 0.4 & 58.2 & 0.5 & 12.5 & 0.5 & 33.6 & 0.3 \\
& 4 & 28.3 & 0.4 & 55.6 & 0.5 & 10.5 & 0.4 & 31.5 & 0.3 \\
& 3 & 23.3 & 0.4 & 43.0 & 0.6 & 6.4 & 0.3 & 24.2 & 0.3 \\
& 2 & 8.1 & 0.2 & 9.3 & 0.3 & 0.0 & 0.0 & 5.8 & 0.1 \\
\hline
\end{tabular}
}
\end{table}
\end{minipage}

}
\end{figure*}

\vspace{4em}

\begin{figure*}[h]
\centering
\makebox[\textwidth][c]{

\begin{minipage}[t]{0.48\textwidth}
\begin{table}[H]
\centering
\parbox[c][3.7\baselineskip][c]{\textwidth}{
\caption{GPTQ vs. NF4 \& LLM.int8() Results}
\label{tab:nf4_llmint8}
}
\resizebox{\textwidth}{!}{
\begin{tabular}{|c|cc|cc|cc|cc|}
\hline
\multirow{2}{*}{\textbf{Model}} & \multicolumn{2}{c|}{\textbf{MMLU-Pro}} & \multicolumn{2}{c|}{\textbf{BBH}} & \multicolumn{2}{c|}{\textbf{MATH}} & \multicolumn{2}{c|}{\textbf{Mean}} \\
\cline{2-9}
& \textbf{\textit{Score}} & \textbf{\textit{Std}} & \textbf{\textit{Score}} & \textbf{\textit{Std}} & \textbf{\textit{Score}} & \textbf{\textit{Std}} & \textbf{\textit{Score}} & \textbf{\textit{Std}} \\
\hline
\hline
LLaMA GPTQ 4bit & 32.0 & 0.4 & 53.3 & 0.5 & 0.2 & 0.1 & 28.5 & 0.2 \\
LLaMA NF4 & 32.0 & 0.4 & 57.0 & 0.5 & 16.8 & 0.5 & 35.2 & 0.3 \\
LLaMA 25\% GPTQ 4bit & 30.5 & 0.4 & 53.4 & 0.5 & 10.5 & 0.5 & 31.5 & 0.3 \\
LLaMA 25\% NF4 & 31.1 & 0.4 & 56.9 & 0.6 & 12.6 & 0.5 & 33.5 & 0.3 \\
\hline
Mistral GPTQ 4bit & 28.3 & 0.4 & 55.6 & 0.5 & 10.5 & 0.4 & 31.5 & 0.3 \\
Mistral NF4 & 29.0 & 0.4 & 57.4 & 0.5 & 12.0 & 0.5 & 32.8 & 0.3 \\
Mistral 25\% GPTQ 4bit & 26.9 & 0.4 & 53.5 & 0.6 & 10.1 & 0.4 & 30.2 & 0.3 \\
Mistral 25\% NF4 & 29.3 & 0.4 & 55.7 & 0.6 & 10.9 & 0.4 & 32.0 & 0.3 \\
\hline
\hline
LLaMA GPTQ 8bit & 35.5 & 0.4 & 62.7 & 0.5 & 18.0 & 0.5 & 38.7 & 0.3 \\
LLaMA LLM.int8() & 34.8 & 0.4 & 62.8 & 0.5 & 18.0 & 0.5 & 38.5 & 0.3 \\
LLaMA 50\% GPTQ 8bit & 20.5 & 0.4 & 40.4 & 0.6 & 3.1 & 0.2 & 21.3 & 0.2 \\
LLaMA 50\% LLM.int8() & 19.8 & 0.4 & 39.9 & 0.6 & 2.7 & 0.2 & 20.8 & 0.2 \\
\hline
Mistral GPTQ 8bit & 30.2 & 0.4 & 58.2 & 0.5 & 12.5 & 0.5 & 33.6 & 0.3 \\
Mistral LLM.int8() & 30.3 & 0.4 & 57.5 & 0.5 & 12.6 & 0.5 & 33.5 & 0.3 \\
Mistral 50\% GPTQ 8bit & 21.7 & 0.4 & 43.8 & 0.5 & 5.0 & 0.3 & 23.5 & 0.2 \\
Mistral 50\% LLM.int8() & 21.6 & 0.4 & 44.0 & 0.5 & 5.1 & 0.3 & 23.6 & 0.2 \\
\hline
\end{tabular}
}
\end{table}
\end{minipage}

\begin{minipage}[t]{0.48\textwidth}
\begin{table}[H]
\centering
\parbox[c][3.7\baselineskip][c]{\textwidth}{
\caption{SparseGPT+GPTQ, Semi-Structured Joint Results}
\label{tab:joint_struct_perf}
}
\resizebox{\textwidth}{!}{
\begin{tabular}{|c|cc|cc|cc|cc|}
\hline
\multirow{2}{*}{\textbf{Model}} & \multicolumn{2}{c|}{\textbf{MMLU-Pro}} & \multicolumn{2}{c|}{\textbf{BBH}} & \multicolumn{2}{c|}{\textbf{MATH}} & \multicolumn{2}{c|}{\textbf{Mean}} \\
\cline{2-9}
& \textbf{\textit{Score}} & \textbf{\textit{Std}} & \textbf{\textit{Score}} & \textbf{\textit{Std}} & \textbf{\textit{Score}} & \textbf{\textit{Std}} & \textbf{\textit{Score}} & \textbf{\textit{Std}} \\
\hline
\hline
LLaMA 3bit & 21.3 & 0.4 & 43.8 & 0.6 & 4.2 & 0.3 & 23.1 & 0.2 \\
LLaMA 25\% 4bit & 30.5 & 0.4 & 53.4 & 0.5 & 10.5 & 0.5 & 31.5 & 0.3 \\
LLaMA 2:8 4bit & 27.7 & 0.4 & 52.9 & 0.6 & 9.6 & 0.4 & 30.1 & 0.3 \\
LLaMA 1:4 4bit & 25.7 & 0.4 & 49.1 & 0.6 & 5.3 & 0.3 & 26.7 & 0.2 \\
\hline
Mistral 3bit & 23.3 & 0.4 & 43.0 & 0.6 & 6.4 & 0.3 & 24.2 & 0.3 \\
Mistral 25\% 4bit & 26.9 & 0.4 & 53.5 & 0.6 & 10.1 & 0.4 & 30.2 & 0.3 \\
Mistral 2:8 4bit & 26.1 & 0.4 & 50.3 & 0.6 & 9.1 & 0.4 & 28.5 & 0.3 \\
Mistral 1:4 4bit & 24.2 & 0.4 & 50.1 & 0.6 & 8.0 & 0.4 & 27.4 & 0.3 \\
\hline
\hline
LLaMA 2bit & 1.9 & 0.1 & 0.1 & 0.0 & 0.0 & 0.0 & 0.6 & 0.0 \\
LLaMA 33.333\% 3bit & 20.3 & 0.4 & 41.5 & 0.6 & 0.8 & 0.1 & 20.9 & 0.2 \\
LLaMA 2:6 3bit & 3.3 & 0.2 & 33.0 & 0.5 & 0.0 & 0.0 & 12.1 & 0.2 \\
LLaMA 1:3 3bit & 11.8 & 0.3 & 32.6 & 0.5 & 0.0 & 0.0 & 14.8 & 0.2 \\
\hline
Mistral 2bit & 8.1 & 0.2 & 9.3 & 0.3 & 0.0 & 0.0 & 5.8 & 0.1 \\
Mistral 33.333\% 3bit & 20.2 & 0.4 & 40.0 & 0.6 & 5.3 & 0.3 & 21.8 & 0.2 \\
Mistral 2:6 3bit & 15.6 & 0.3 & 33.7 & 0.5 & 2.3 & 0.2 & 17.2 & 0.2 \\
Mistral 1:3 3bit & 15.0 & 0.3 & 34.2 & 0.5 & 2.8 & 0.2 & 17.4 & 0.2 \\
\hline
\end{tabular}
}
\end{table}
\end{minipage}

}
\end{figure*}

\vspace{4em}

\begin{figure*}[h]
\centering
\makebox[\textwidth][c]{

\begin{minipage}[t]{0.48\textwidth}
\begin{table}[H]
\centering
\parbox[c][3.7\baselineskip][c]{\textwidth}{
\caption{SparseGPT, Semi-Structured Pruning-Only Results}
\label{tab:sparsegpt_struct_perf}
}
\resizebox{\textwidth}{!}{
\begin{tabular}{|c|c|cc|cc|cc|cc|}
\hline
\multirow{2}{*}{\textbf{Model}} & \multirow{2}{*}{\textbf{Sparsity}} & \multicolumn{2}{c|}{\textbf{MMLU-Pro}} & \multicolumn{2}{c|}{\textbf{BBH}} & \multicolumn{2}{c|}{\textbf{MATH}} & \multicolumn{2}{c|}{\textbf{Mean}} \\
\cline{3-10}
& & \textbf{\textit{Score}} & \textbf{\textit{Std}} & \textbf{\textit{Score}} & \textbf{\textit{Std}} & \textbf{\textit{Score}} & \textbf{\textit{Std}} & \textbf{\textit{Score}} & \textbf{\textit{Std}} \\
\hline
\multirow{7}{*}{LLaMA 8B}
& 0\% & 35.4 & 0.4 & 62.3 & 0.5 & 17.7 & 0.5 & 38.5 & 0.3 \\
\cline{2-10}
& 25\% & 33.3 & 0.4 & 60.2 & 0.5 & 14.9 & 0.5 & 36.2 & 0.3 \\
& 2:8 & 31.0 & 0.4 & 57.3 & 0.5 & 13.6 & 0.5 & 34.0 & 0.3 \\
& 1:4 & 28.4 & 0.4 & 53.8 & 0.6 & 10.1 & 0.4 & 30.8 & 0.3 \\
\cline{2-10}
& 33.333\% & 30.2 & 0.4 & 55.2 & 0.5 & 11.5 & 0.4 & 32.3 & 0.3 \\
& 2:6 & 21.2 & 0.4 & 45.3 & 0.6 & 2.8 & 0.2 & 23.1 & 0.2 \\
& 1:3 & 19.2 & 0.4 & 43.3 & 0.6 & 2.5 & 0.2 & 21.7 & 0.2 \\
\hline
\multirow{7}{*}{Mistral 7B} %10
& 0\% & 30.3 & 0.4 & 58.0 & 0.5 & 12.8 & 0.5 & 33.7 & 0.3 \\
\cline{2-10}
& 25\% & 29.4 & 0.4 & 55.7 & 0.6 & 10.9 & 0.4 & 32.0 & 0.3 \\
& 2:8 & 27.9 & 0.4 & 53.3 & 0.6 & 9.9 & 0.4 & 30.4 & 0.3 \\
& 1:4 & 27.0 & 0.4 & 52.1 & 0.6 & 9.2 & 0.4 & 29.4 & 0.3 \\
\cline{2-10}
& 33.333\% & 27.9 & 0.4 & 53.9 & 0.6 & 10.0 & 0.4 & 30.6 & 0.3 \\
& 2:6 & 21.8 & 0.4 & 44.3 & 0.5 & 5.9 & 0.3 & 24.0 & 0.2 \\
& 1:3 & 21.7 & 0.4 & 44.0 & 0.6 & 5.8 & 0.3 & 23.8 & 0.2 \\
%\cline{2-10}
%& 50\% & 21.6 & 0.4 & 44.0 & 0.5 & 5.1 & 0.3 & 23.6 & 0.2 \\
%& 4:8 & 15.8 & 0.3 & 36.6 & 0.6 & 2.3 & 0.2 & 18.2 & 0.2 \\
%& 2:4 & 13.5 & 0.3 & 32.3 & 0.5 & 1.4 & 0.2 & 15.8 & 0.2 \\
\hline
\end{tabular}
}
\end{table}
\end{minipage}

\begin{minipage}[t]{0.48\textwidth}
\begin{table}[H]
\centering
\parbox[c][3.7\baselineskip][c]{\textwidth}{
\caption{SparseGPT+GPTQ, Unstructured Joint Results}
\label{tab:joint_unstruct_perf}
}
\resizebox{\textwidth}{!}{
\begin{tabular}{|c|cc|cc|cc|cc|}
\hline
\multirow{2}{*}{\textbf{Model}} & \multicolumn{2}{c|}{\textbf{MMLU-Pro}} & \multicolumn{2}{c|}{\textbf{BBH}} & \multicolumn{2}{c|}{\textbf{MATH}} & \multicolumn{2}{c|}{\textbf{Mean}} \\
\cline{2-9}
& \textbf{\textit{Score}} & \textbf{\textit{Std}} & \textbf{\textit{Score}} & \textbf{\textit{Std}} & \textbf{\textit{Score}} & \textbf{\textit{Std}} & \textbf{\textit{Score}} & \textbf{\textit{Std}}\\
\hline
\hline
LLaMA 4bit & 32.0 & 0.4 & 53.3 & 0.5 & 0.2 & 0.1 & 28.5 & 0.2 \\
LLaMA 50\% 8bit & 20.5 & 0.4 & 40.4 & 0.6 & 3.1 & 0.2 & 21.3 & 0.2 \\
\hline
Mistral 4bit & 28.3 & 0.4 & 55.6 & 0.5 & 10.5 & 0.4 & 31.5 & 0.3 \\
Mistral 50\% 8bit & 21.7 & 0.4 & 43.8 & 0.5 & 5.0 & 0.3 & 23.5 & 0.2 \\
\hline
\hline
LLaMA 3bit & 21.3 & 0.4 & 43.8 & 0.6 & 4.2 & 0.3 & 23.1 & 0.2 \\
LLaMA 25\% 4bit & 30.5 & 0.4 & 53.4 & 0.5 & 10.5 & 0.5 & 31.5 & 0.3 \\
\hline
Mistral 3bit & 23.3 & 0.4 & 43.0 & 0.6 & 6.4 & 0.3 & 24.2 & 0.3 \\
Mistral 25\% 4bit & 26.9 & 0.4 & 53.5 & 0.6 & 10.1 & 0.4 & 30.2 & 0.3 \\
\hline
\hline
LLaMA 2bit & 1.9 & 0.1 & 0.1 & 0.0 & 0.0 & 0.0 & 0.6 & 0.0 \\
LLaMA 33.333\% 3bit & 20.3 & 0.4 & 41.5 & 0.6 & 0.8 & 0.1 & 20.9 & 0.2 \\
\hline
Mistral 2bit & 8.1 & 0.2 & 9.3 & 0.3 & 0.0 & 0.0 & 5.8 & 0.1 \\
Mistral 33.333\% 3bit & 20.2 & 0.4 & 40.0 & 0.6 & 5.3 & 0.3 & 21.8 & 0.2 \\
\hline
\end{tabular}
}
\end{table}
\end{minipage}

}
\end{figure*}

\end{document}